\documentclass{article}
\def\arxivversion{}
\usepackage{listings}
\usepackage[table]{xcolor}
\usepackage{graphicx}
\graphicspath{{./}}
\usepackage{subcaption}
\usepackage[utf8]{inputenc}
\usepackage{microtype}
\usepackage{booktabs}
\usepackage{hyperref}
\usepackage[most]{tcolorbox}
\usepackage{tabularx}
\usepackage{multirow}
\usepackage{makecell}
\usepackage{float}
\usepackage[htt]{hyphenat}
\usepackage{array}
\newcolumntype{L}[1]{>{\raggedright\arraybackslash}m{#1}}
\usepackage{amsmath}
\usepackage{amsfonts}
\usepackage{amssymb}
\usepackage{url}
\usepackage{xurl}   
\usepackage{balance}
\usepackage{tikz}
\usepackage{etoolbox}
\usepackage{algorithm}
\usepackage{multirow}
\usepackage{booktabs}

\usepackage{xspace}
\usepackage{textcomp}
\usepackage{calc}
\usepackage{lineno}
\definecolor{codegray}{rgb}{0.95,0.95,0.95}
\definecolor{RowBlue}{HTML}{ECF4FF}
\definecolor{RowGreen}{HTML}{E8F7EC}
\definecolor{RowGray}{HTML}{F4F4F4}
\definecolor{RowOrange}{HTML}{FFF1E0}
\definecolor{CellYellow}{HTML}{FFFFE0}

\lstdefinestyle{cppstyle}{
    language=C++,
    basicstyle=\ttfamily\scriptsize,
    numbers=left,
    numberstyle=\tiny,
    stepnumber=1,
    numbersep=6pt,
    breaklines=true,
    breakatwhitespace=false,
    showstringspaces=false,
    tabsize=2,
    frame=none,
    keywordstyle=\color{blue},
    commentstyle=\color{gray},
    stringstyle=\color{teal}
}

\lstdefinestyle{cppstyle}{
    language=C++,
    backgroundcolor=\color{codegray},
    basicstyle=\ttfamily\scriptsize,
    numbers=left,
    numberstyle=\tiny,
    stepnumber=1,
    numbersep=6pt,
    breaklines=true,
    breakatwhitespace=false,
    columns=fullflexible,
    keepspaces=true,
    showstringspaces=false,
    tabsize=2,
    frame=none,
    aboveskip=0pt,
    belowskip=0pt,
    lineskip=0pt,
    keywordstyle=\color{blue},
    commentstyle=\color{gray},
    stringstyle=\color{teal}
}

\newcommand{\highlightyellow}[1]{%
  \leavevmode
  \setlength{\fboxsep}{0pt}%
  \colorbox{yellow!20}{\hbox to \dimexpr\linewidth-\leftmargin\relax{\strut #1\hfill}}%
}
\newcommand{\highlightblue}[1]{%
  \leavevmode
  \setlength{\fboxsep}{0pt}%
  \colorbox{blue!10}{\hbox to \dimexpr\linewidth-\leftmargin\relax{\strut #1\hfill}}%
}
\newcommand{\mem}{\mathcal{K}}
\newcommand{\KB}{\textit{Knowledge Base}\xspace}
\newcommand{\sys}{\textsc{ModelBlaster}\xspace}

\usepackage{mlsys2025}
\usepackage[capitalize,noabbrev]{cleveref}
\mlsystitlerunning{ModelBlaster}

\makeatletter
\renewcommand{\Notice@String}{}
\AtBeginDocument{\ClearShipoutPicture}
\AtBeginDocument{%
  \setlength{\footskip}{18pt}
  \pagestyle{fancy}
  \fancyhf{}%
  \fancyfoot[C]{\thepage}%
  \renewcommand{\headrulewidth}{1pt}%
}
\makeatother
\hypersetup{pdfauthor={Kris Shengjun Dong, Dima Nikiforov, Sahil Modi, Sana Damani, Edward Lin, Siva Kumar Sastry Hari, Christos Kozyrakis}}

\begin{document}
\twocolumn[
\mlsystitle{ModelBlaster: Closing the Performance Gap through Learning-Augmented Agents Across Models, Kernels and Hardware Backends}
\centering{Kris Shengjun Dong$^{1,2}$, Dima Nikiforov$^{2}$, Sahil Modi$^{1}$, Sana Damani$^{1}$,\\
Edward Lin$^{1}$, Siva Kumar Sastry Hari$^{1}$, Christos Kozyrakis$^{1}$\\
$^{1}$NVIDIA, $^{2}$University of California, Berkeley}
\vskip 0.3in
\begin{abstract}
Optimizing accelerator code across hardware architectures is difficult. ML compilers typically search predefined spaces of schedules and micro-kernels. Large language model (LLM)-guided kernel optimization can explore a broader design space, including alternative algorithms, operator decompositions, and hardware-specific techniques, achieving better performance than ML compilers on many problems. However, existing LLM agents generate solutions that are tied to a single programming interface, vendor toolchain, or accelerator family, and target kernel-level optimizations that don't leverage model-level transforms. Additionally, most of the experience gathered on one hardware target cannot be reused on another.

We present \sys, a Memory-Augmented In-Context Reinforcement Learning (MAIC-RL) framework that optimizes data-parallel kernels across heterogeneous SIMD and SIMT accelerator targets. Its mutable policy is a persistent knowledge base, updated at inference time by Actor--Critic Textual Gradient Descent (AC-TGD). Its structured \KB maps performance states to actions, outcomes, and constraints, so experience gathered on one kernel, model, or hardware target can be reused on another. A model-level optimizer extends the kernel agent to whole models, so kernel-level knowledge can inform model-level decisions. The same loop and schema serve CUDA on NVIDIA GPUs, OpenCL on Qualcomm GPUs, and RVV intrinsics on RISC-V vector cores. 

Leveraging the cross-target accumulated knowledge, \sys achieves geometric-mean speedups of 1.85$\times$, 5.58$\times$, and 5.18$\times$ on KernelBench Levels~1--3 on an L40S GPU, and 1.69$\times$ on SOL-ExecBench Levels~1 and 2 on a B200 GPU, comparing to PyTorch. On Qualcomm Adreno GPUs it reaches 5.8$\times$ over the vendor SDK and 1.7$\times$ over Apache TVM, and on RISC-V vector cores 2.7$\times$ over ExecuTorch with XNNPACK across 70 KernelBench problems.

\end{abstract}

]

\section{Introduction}\label{sec:introduction}

The demand for machine-learning (ML) applications has exploded, and a 1\% efficiency gain is now worth hundreds of millions of dollars a year in electricity alone~\cite{eia2025electricity,iea2025energyai,deloitte2026tmt}.
Kernel optimization is a major lever on that efficiency. On H100 GPUs, software optimizations including GEMM and attention kernel optimizations and kernel fusions raised Llama 2 70B throughput by up to 1.5$\times$ in one year~\cite{eassa2025mlperf}, and a single attention kernel redesign gave GPT-2 a 3$\times$ end-to-end speedup~\cite{dao2022flashattention}.

Capturing these gains is hard because hardware and models keep changing. New deep neural network (DNN) architectures and dynamic workloads invalidate tuned kernels, so software stacks must maintain specialized kernel libraries~\cite{sevilla2022compute,chen2025evolutionkernels}, which takes substantial engineering effort and delays the use of new hardware capabilities~\cite{ktt2020benchmark}.

ML compilers automate the translation from models to kernels. Many integrate graph optimization and code generation through shared dialects and feedback passes, with autotuners such as Ansor and XLA's XTAT adding measured feedback~\cite{zheng2020ansor,mb_xtat2021}. However, \emph{what a compiler can find is limited by what it can express.} Halide and TVM are constrained by predefined search spaces, scheduling primitives, and handcrafted rules~\cite{mb_kernelgensurvey2026}, and they optimize inside a fixed space defined by human-crafted templates~\cite{mb_flashinferbench2026}. Autotuners search numeric parameters such as tile sizes inside a fixed kernel structure, so they cannot change a kernel's algorithm, data types, or fusion~\cite{mb_xeforge2026}, and they struggle when a gain needs a structural code edit~\cite{mb_kernelbrain2026}. A tile-level fusion that overlaps activation work with matrix multiplication shows the cost: \texttt{torch.compile} cannot replicate it and is 3 to 7 times slower than the hand-crafted kernels~\cite{mb_tilelevel2026}. Even within the expressible space, stochastic search is sample-inefficient and misses transformations that pay off only in combination~\cite{mb_reasoningcompiler2025}, and hand-tuned microkernels perform better in mixed-precision calculations~\cite{mb_ireervv2025}.

Large language model (LLM) agents offer a way around those abstractions. They can customize, check, and profile kernels during optimization, including restructurings that no predefined schedule family contains~\cite{Lange2025AICUDAEngineer,novikov2025alphaevolve,mb_accelopt2026}, and they can outperform ML compilers on individual kernels~\cite{mb_kernelopt2026,mb_nagaitsev2025,gim2025pie,lin2025eco,openevolve2025,Baronio2025Kevin32B,Cummins2025LLMCompiler,Taneja2025LlmVectorizer}. However, most existing agents optimize each kernel in isolation. They sample candidate implementations, observe execution feedback, and revise the code, but the lessons stay inside the task: when it ends, most of the experience is discarded, so the next kernel or target starts with little prior knowledge~\cite{mb_rethinkingagentic2026,mb_kernelgensurvey2026}. Many of these agents are also tied to a single programming interface, vendor toolchain, or accelerator family~\cite{mb_kernelgensurvey2026,mb_kforge2026}.

This limits heterogeneous optimization, where the agent must reason over backend-independent concepts such as locality, occupancy, tiling, vectorization, and fusion while emitting backend-specific code and reading backend-specific profiling signals~\cite{mb_kernelgensurvey2026,mb_kforge2026}. Search, prompt-optimization, and evolutionary methods keep some optimization signal but generally lack an executable, persistent memory of profile-grounded state, action, and outcome evidence that transfers across kernels and targets. These agents also work on one kernel at a time, since a whole model does not fit in one prompt~\cite{mb_kernelgensurvey2026,mb_llm4llm2026,ouyang2024kernelbench}.

\textbf{\sys} introduces a \emph{Memory-Augmented In-Context Reinforcement Learning} (MAIC-RL) framework for cross-task code optimization. Instead of updating model weights, it maintains a persistent \KB of performance states, optimization actions, target constraints, textual critiques, and measured outcomes. An actor generates memory-guided transformations, while a critic evaluates correctness, performance, and profiling feedback to produce text-based rewards that update the \KB with successful and failed experiences.

Extending from kernels to whole models, \sys extracts a PyTorch computation graph, proposes transformations through versioned hints, and dispatches selected partitions to the kernel agent. Graph changes are retained only when verified, optimized kernels improve measured whole-model performance. This feedback is necessary because dynamically generated kernels can alter the relative costs of fusion and partitioning decisions.

We evaluate \sys across heterogeneous languages, ISAs, and hardware: CUDA C++ on NVIDIA GPUs, OpenCL C on Qualcomm GPUs, and RVV intrinsics on RISC-V vector cores. The shared optimization methodology requires no target-specific training data, supporting low-resource architectures. On B200, \sys achieves geometric-mean speedups of 1.85$\times$, 5.58$\times$, and 5.18$\times$ over PyTorch on KernelBench Levels~1--3, respectively, and 1.69$\times$ on SOL-ExecBench~\cite{lin2026solexecbench} Levels~1--2. It also achieves 5.8$\times$ over QNN and 1.7$\times$ over Apache TVM on Qualcomm Adreno, and 2.7$\times$ over ExecuTorch/XNNPACK on RISC-V vector cores.

The contributions of this paper are:\vspace{-7pt}
\begin{enumerate}\setlength{\itemsep}{0pt}\setlength{\parsep}{0pt}\setlength{\parskip}{-2pt}\setlength{\topsep}{0pt}\setlength{\partopsep}{0pt}
\item \textbf{One framework for graph and kernel optimization, with feedback from kernels to graph.} Because AI-generated kernels change the implementation space during the search, graph rewrites create kernel tasks, and the verified kernels that come back can confirm or reverse decisions made from earlier cost estimates, within a fixed search budget.
\item \textbf{One harness across heterogeneous targets.} The same data structures, \KB schema, and method run on RISC-V vector cores, Qualcomm mobile GPUs, and NVIDIA GPUs, with only lowering, compilation, validation, and profiling specific to each backend.
\item \textbf{An actor and critic agent architecture for policy improvement.} Scalar execution outcomes and natural-language critiques, derived from compilation, correctness, profiling, and performance-gap evidence, act as textual gradients that update memory and guide later actions.
\item \textbf{A reusable long-term memory.} A structured database of what was tried and measured, including failures, is indexed by hardware-profile state and reusable across tasks, targets, and languages. Entries learned on one vendor's hardware may not help on another vendor's or on a different instruction set.
\end{enumerate}
\vspace{-7pt}
The full code base, including the agentic pipeline, baseline kernels, initialized databases, and test harnesses, has been open-sourced.\ifdefined\arxivversion\footnote{The code base is released at \mbox{\url{https://github.com/ucb-bar/ModelBlaster}}.}\else\footnote{An anonymous mirror of the full code base is released at \url{https://anonymous.4open.science/r/ModelBlaster/} to respect the double-blind procedure.}\fi
\section{Background and Related Work}\label{sec:related}

\usetikzlibrary{decorations.pathreplacing,calc}
\newcommand{\tedge}[2]{\draw let \p1=(#1.south), \p2=(#2.north) in (\x1,\y1) -- (\x1,{(\y1+\y2)/2}) -| (\x2,\y2);}
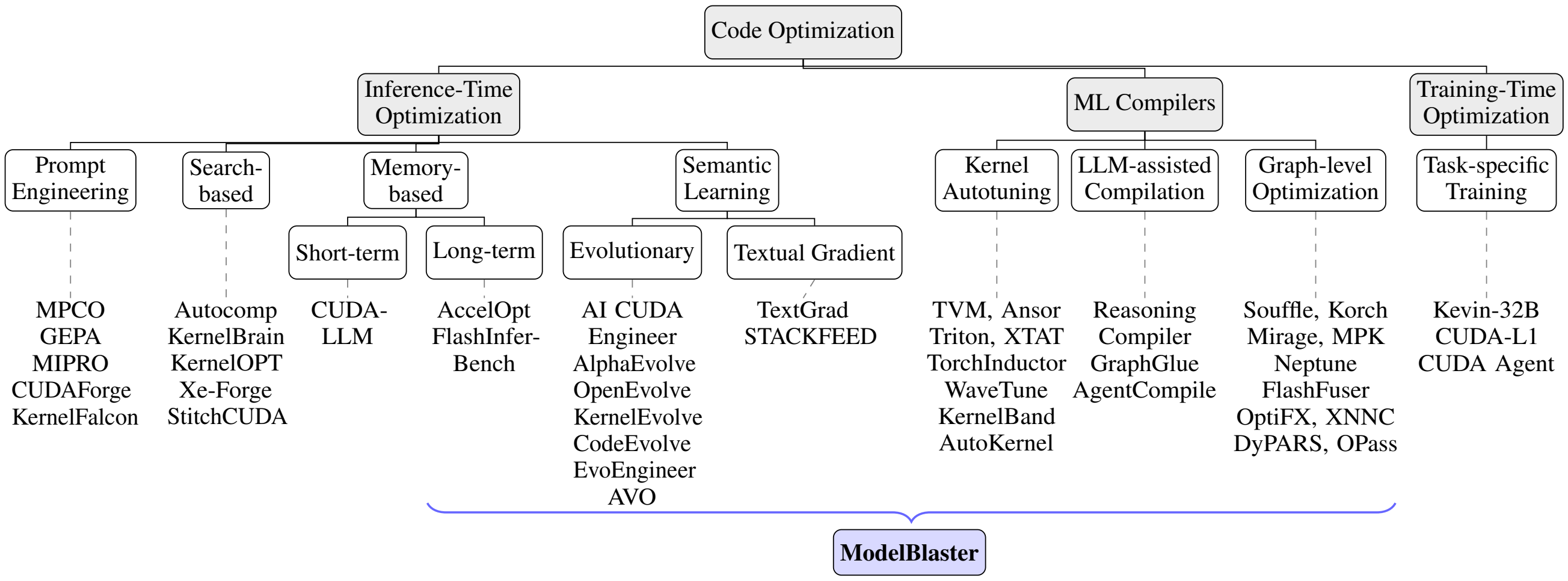
\begin{figure*}[t]
\centering
\resizebox{\textwidth}{!}{%
\begin{tikzpicture}[x=1cm,y=1cm,
  n/.style={draw,rounded corners=3pt,align=center,font=\footnotesize,inner sep=2.5pt,minimum height=0.7cm},
  top/.style={n,fill=gray!15},
  ours/.style={draw,rounded corners=3pt,align=center,font=\footnotesize\bfseries,inner sep=2.5pt,fill=blue!15,minimum height=0.6cm},
  ex/.style={font=\footnotesize,align=center,text width=1.55cm,inner sep=1pt,anchor=north},
  dl/.style={dashed,gray}]
  \node[top] (root) at (10.3,0) {Code Optimization};
  \node[top] (inf) at (5.5,-0.95) {Inference-Time\\Optimization};
  \node[top] (cmp) at (14.8,-0.95) {ML Compilers};
  \node[top] (trn) at (19.3,-0.95) {Training-Time\\Optimization};
  \node[n] (prm) at (0.65,-1.95) {Prompt\\Engineering};
  \node[n] (sea) at (2.7,-1.95) {Search-\\based};
  \node[n] (mem) at (5.2,-1.95) {Memory-\\based};
  \node[n] (sem) at (9.3,-1.95) {Semantic\\Learning};
  \node[n] (aut) at (12.85,-1.95) {Kernel\\Autotuning};
  \node[n] (llc) at (14.8,-1.95) {LLM-assisted\\Compilation};
  \node[n] (grp) at (17.05,-1.95) {Graph-level\\Optimization};
  \node[n] (sft) at (19.3,-1.95) {Task-specific\\Training};
  \node[n] (shr) at (4.3,-2.9) {Short-term};
  \node[n] (lng) at (6.1,-2.9) {Long-term};
  \node[n] (evo) at (8.05,-2.9) {Evolutionary};
  \node[n] (tgd) at (10.45,-2.9) {Textual Gradient};
  \foreach \c in {inf,cmp,trn} {\tedge{root}{\c}}
  \foreach \c in {prm,sea,mem,sem} {\tedge{inf}{\c}}
  \foreach \c in {aut,llc,grp} {\tedge{cmp}{\c}}
  \tedge{trn}{sft}
  \foreach \c in {shr,lng} {\tedge{mem}{\c}}
  \foreach \c in {evo,tgd} {\tedge{sem}{\c}}
  \node[ex] (e1) at (0.65,-3.5)  {MPCO\\GEPA\\MIPRO\\CUDAForge\\KernelFalcon};
  \node[ex] (e2) at (2.7,-3.5) {Autocomp\\KernelBrain\\KernelOPT\\Xe-Forge\\StitchCUDA};
  \node[ex] (e3) at (4.3,-3.5)  {CUDA-LLM};
  \node[ex] (e4) at (6.1,-3.5) {AccelOpt\\FlashInfer-\\Bench};
  \node[ex] (e5) at (8.05,-3.5)  {AI CUDA\\Engineer\\AlphaEvolve\\OpenEvolve\\KernelEvolve\\CodeEvolve\\EvoEngineer\\AVO};
  \node[ex] (e6) at (10.3,-3.5) {TextGrad\\STACKFEED};
  \node[ex,text width=1.8cm] (e7) at (19.3,-3.5) {Kevin-32B\\CUDA-L1\\CUDA Agent};
  \node[ex,text width=2.0cm] (e9) at (12.85,-3.5) {TVM, Ansor\\Triton, XTAT\\TorchInductor\\WaveTune\\KernelBand\\AutoKernel};
  \node[ex,text width=1.9cm] (e10) at (14.8,-3.5) {Reasoning Compiler\\GraphGlue\\AgentCompile};
  \node[ex,text width=2.4cm] (e11) at (17.05,-3.5) {Souffle, Korch\\Mirage, MPK\\Neptune\\FlashFuser\\OptiFX, XNNC\\DyPARS, OPass};
  \foreach \a/\b in {prm/e1,sea/e2,shr/e3,lng/e4,evo/e5,tgd/e6,sft/e7,aut/e9,llc/e10,grp/e11} {\draw[dl] (\a.south) -- (\b.north);}
  \draw[decorate,decoration={brace,mirror,amplitude=7pt},thick,blue!60] (5.35,-6.2) -- (18.1,-6.2);
  \node[ours] (o) at (11.7,-6.85) {ModelBlaster};
\end{tikzpicture}}
\caption{Taxonomy of approaches to ML code optimization. \sys, at the bottom, combines long-term memory, textual-gradient learning, and the kernel-level and graph-level functions of ML compilers.}
\label{fig:taxonomy}
\end{figure*}

\subsection{Compilers and Autotuning}\label{subsec:compilers}
\textbf{Kernel autotuning.} Conventional ML compilers apply fusion and layout rules, guided by heuristics and cost models, then instantiate each fused operator from micro-kernel specifications, library calls, or schedule templates whose parameters an autotuner searches with measured feedback~\cite{chen2018tvm,zheng2020ansor,mb_xtat2021}. A feedback loop exists, but the search space is fixed in advance by the abstraction. Kernel-level autotuners such as TorchInductor with \texttt{max-autotune}, TVM, Ansor, and Triton profile several template-based implementations and keep the fastest~\cite{ansel2024pytorch2,chen2018tvm,zheng2020ansor,tillet2019triton}. LLM-era frameworks add analytical models and search: WaveTune~\cite{zhang2026wavetune} uses wave-aware performance models for tuning on fixed kernel sets, KernelBand~\cite{ran2025kernelband} uses hierarchical bandits with profiling-aware clustering, and AutoKernel~\cite{jaber2026autokernel} uses edit-and-benchmark loops with curated playbooks. These target deep single-backend tuning, while \sys targets broader operator coverage and cross-target reuse through a persistent memory of states and actions.

\textbf{LLM-assisted compilation.} Reasoning Compiler~\cite{mb_reasoningcompiler2025} pairs LLM-proposed transformations with Monte Carlo tree search to reduce the sample inefficiency of stochastic compiler search, and GraphGlue~\cite{mb_graphglue2025} uses an LLM to repair graph-capture breaks so the compiler sees more of the program. AgentCompile~\cite{mb_agentcompile2026} combines compiler-derived region summaries with LLM-generated kernels for transformer inference on CUDA and is the closest of these to our setting. LLM assistance inside a compiler is thus established.

\textbf{Graph-level optimization.} Fusion, splitting, tiling, and layout changes pay off only given the kernels available for the target. Souffle~\cite{mb_souffle2024} splits a global tensor-dependency graph into subprograms across operator boundaries, Korch~\cite{mb_korch2024} assigns operator primitives to kernels with binary linear programming, and Mirage~\cite{mb_mirage2025} searches algebraic and scheduling transformations together with custom kernel generation. Mirage is the closest to our view that graph and kernel spaces are coupled, but it uses a different superoptimization and verification approach. MPK~\cite{mb_mpk2026} runs inference as one persistent mega-kernel, Neptune~\cite{mb_neptune2026} fuses reduction-heavy operators by rewriting dependencies, and FlashFuser~\cite{mb_flashfuser2026} fuses operators whose intermediates exceed shared memory using H100 distributed shared memory. OptiFX~\cite{mb_optifx2025} and XNNC~\cite{mb_xnnc2025} search for aggressive fusion, DyPARS~\cite{mb_dypars2026} finds dynamic-shape graph variants with Pareto-aware tree search, and OPass~\cite{mb_opass2026} orders TVM graph passes to cut memory. These systems show that sophisticated fusion and joint graph and kernel search are not new by themselves, and they rely on predefined transformation rules, templates, or implementation spaces. With a fixed kernel library, graph choices can be made before specialization. When agents generate the kernels, the set of implementations grows during the search, so a new kernel can reverse an earlier graph decision. \sys connects a graph stage with an LLM-generated kernel search checked by execution, keeps experience across tasks and targets, and lets an external graph compiler take the graph stage.

\subsection{Agentic Solutions}\label{subsec:kernel-related}
Agentic systems for code optimization are broadly categorized as shown in Figure~\ref{fig:taxonomy}. We focus on inference-time methods that avoid weight updates and use frontier LLMs. They differ mainly in what adapts across attempts: prompt engineering keeps hand-written heuristics, search keeps candidate trajectories within a task, memory keeps records of earlier attempts, and semantic learning keeps candidate programs (evolutionary) or textual feedback (textual gradients).

\textbf{Prompt engineering.} Early LLM-based optimization systems use fixed prompts with manually engineered heuristics~\cite{zhang2025cudaforge,wang2025kernelfalcon}, and prompt optimizers search over instructions: MIPRO~\cite{opsahl-ong2024optimizing}, GEPA with reflective prompt evolution and Pareto-based selection~\cite{agrawal2025gepa}, and MPCO with meta-prompting that generates task-specific prompts for each LLM~\cite{mb_mpco2025}. Their behavior is mostly fixed by the prompt, so a new architecture, toolchain, or workload often needs a redesigned prompt, which limits reuse of experience gathered during inference.

\textbf{Search-based optimization.} Search-based systems generate several candidate kernels and keep high-performing variants using compilation, testing, and timing feedback~\cite{hong2025autocompllmdrivencodeoptimization,mb_kernelbrain2026,mb_kernelopt2026,mb_xeforge2026}. Pruning and budget-aware, coarse-to-fine search improve sample efficiency over exhaustive enumeration. StitchCUDA~\cite{li2026stitchcuda} organizes the search as a multi-agent loop of a planner, a coder, and a verifier that profiles with Nsys and NCU, and it can run with a closed model. The learning is usually task-local: each new kernel starts another search, and few bottleneck-level lessons transfer between tasks.

\textbf{Memory-based optimization.} Short-term memory keeps the history of one task. CUDA-LLM~\cite{chen2025cuda_llm} refines kernels over rounds of compilation, correctness, and timing feedback, and the history is discarded when the task ends. Long-term memory carries records across tasks: AccelOpt~\cite{mb_accelopt2026} keeps a memory of slow and fast kernel candidates and uses beam search for AI-accelerator kernels, and FlashInfer-Bench~\cite{mb_flashinferbench2026} connects agent-generated kernels, benchmarking, and deployment into serving engines. \sys indexes memory by hardware-profile state (Section~\ref{subsec:profiling}).

\textbf{Evolutionary coding agents.} Systems such as AI CUDA Engineer~\cite{Lange2025AICUDAEngineer,lange2025robustkbench}, KernelEvolve~\cite{liao2025kernelevolve}, AlphaEvolve~\cite{novikov2025alphaevolve}, OpenEvolve~\cite{openevolve2025}, CodeEvolve~\cite{assumpcao2025codeevolve}, and EvoEngineer~\cite{guo2025evoengineer} keep populations of candidate programs and improve them through LLM-generated variation and evaluator feedback. AVO~\cite{mb_avo2026} replaces fixed mutation and crossover with a coding agent that uses the kernel lineage, a knowledge base, and execution feedback. Their memory is usually full programs or elite candidates, which help reuse implementations but can hide why an optimization worked. \sys instead stores compact, bottleneck-level records of state, action, and outcome, including negative evidence, so retrieval follows the performance state and not only source-code similarity.

\textbf{Textual gradients.} TextGrad~\cite{yuksekgonul2024textgrad} uses natural-language feedback as an analogue of gradients over textual variables, and STACKFEED~\cite{kirtania2025stackfeed} uses text as a gradient in an actor and critic loop but edits retrieval knowledge for question answering. These methods mainly adapt to one task and often must be optimized again when the task distribution, toolchain, or environment changes, which limits transfer across systems tasks where one optimization concept appears under different languages, backends, and profiling interfaces. \sys builds on the textual-gradient idea and changes the target: textual gradients update a persistent \KB of state, action, and outcome records that later tasks can retrieve and reuse. Our ablations separate the effect of memory from the effect of profiling and from a minimal agent (Appendix~\ref{app:ablations}).


\textbf{Kernel agents and whole models.} LLM-guided optimization removes the restriction to a predefined space. An agent can propose algorithmic restructurings, different operator decompositions, and hardware-specific techniques that no predefined dimension covers~\cite{mb_kernelgensurvey2026}. Most agentic systems, however, work on one kernel or one small task at a time and treat a whole model as a set of independent kernel tasks, so no stage decides which partitions deserve a generated kernel. Generated kernels also have limited reach where libraries already sit near the hardware limit, such as the GEMM and attention kernels that dominate transformer runtime~\cite{dao2022flashattention,mb_realworkload2026}, so a model-level view has to decide where a generated kernel is worth writing. None of these kernel agents optimizes the model graph. \sys adds a model-level optimizer that proposes graph transformations, sends the resulting partitions to the kernel agent, and keeps a change only if the verified kernels lower the measured whole-model cost.

\section{System Design}\label{sec:method}

\sys optimizes accelerator code with a kernel-level agent and extends it to whole models with a model-level optimizer. Both share a persistent \KB and a harness that compiles, validates, and profiles (Figure~\ref{fig:architecture}). Section~\ref{subsec:kernel-level} describes the kernel agent, Section~\ref{subsec:arch} its extension to whole models, and Section~\ref{subsec:modular} the interfaces between the parts.

\begin{figure*}[h]
\centering

\begin{minipage}[t]{0.53\textwidth}
\footnotesize
\captionof{algorithm}{REINFORCE}
\label{alg:PG}
\begin{algorithmic}[1]
\STATE \textbf{Input:} Initial policy params $\theta_0$, env.~$\mathcal{E}$, policy $\pi_\theta$
\STATE Initialize replay buffer $\mathcal{D} \gets \emptyset$
\FOR{iteration $k = 0, 1, 2, \ldots$}
    \STATE Sample initial state $s_0 \sim \mathcal{E}$
    \STATE Initialize trajectory $\tau \gets []$
    \STATE $s \gets s_0$
    \FOR{step $t = 0$ to $T$}
        \STATE \highlightblue{Sample action $a_t \sim \pi_{\theta_k}(a_t \mid s_t)$}
        \STATE \highlightblue{Append $(s_t, a_t)$ to trajectory $\tau$}
        \STATE \highlightblue{Set $s_{t+1} \gets \texttt{EnvStep}(s_t, a_t)$}
        \STATE \highlightblue{Set $r_t = R(s_t, a_t)$}
    \ENDFOR
    \STATE Evaluate return $R(\tau) = \sum_t r_t$
    \STATE \highlightyellow{Store $(\tau, R(\tau))$ in buffer $\mathcal{D}$}
    \STATE \highlightyellow{Estimate policy gradient:}
    \STATE \highlightyellow{$\nabla_\theta J(\theta_k) \gets \frac{1}{|\mathcal{D}|} \sum_{(\tau, R)} \sum_t \nabla_\theta \log \pi_\theta(a_t \mid s_t) R(\tau)$}
    \STATE \highlightyellow{Update policy: $\theta_{k+1} \gets \theta_k + \alpha \nabla_\theta J(\theta_k)$}
\ENDFOR
\end{algorithmic}
\end{minipage}
\hfill
\begin{minipage}[t]{0.46\textwidth}
\footnotesize
\captionof{algorithm}{LLM-Based Policy Opt.~via Reward-Guided Rollouts}
\label{alg:ICRL}
\begin{algorithmic}[1]
\STATE \textbf{Input:} Initial \KB$\theta_0$, env.~$\mathcal{E}$, LLM policy $\pi_\theta$
\STATE Initialize replay buffer $\mathcal{D} \gets \emptyset$
\FOR{iteration $k = 0, 1, 2, \ldots$}
    \STATE Sample code task $s_0 \sim \mathcal{E}$ \COMMENT{Initial code}
    \STATE Initialize rollout trajectory $\tau \gets []$
    \STATE $s \gets s_0$
    \FOR{step $t = 0$ to $T$}
        \STATE \highlightblue{Generate optimized code $a_t \sim \pi_{\theta_k}(a_t \mid s_t)$}
        \STATE \highlightblue{Append $(s_t, a_t)$ to trajectory $\tau$}
        \STATE \highlightblue{Set $s_{t+1} \gets a_t$ \COMMENT{Set new code as state}}
        \STATE \highlightblue{Set $r_t = R(s_{t+1})$ \COMMENT{Profile-based reward}}
    \ENDFOR
    \STATE Evaluate total reward $R(\tau)$
    \STATE \highlightyellow{Store $(\tau, R(\tau))$ in buffer $\mathcal{D}$}
    \STATE \highlightyellow{$g_k \gets \texttt{PolicyEvaluation}(\mathcal{D}, \theta_k)$}
    \STATE \highlightyellow{$p_k \gets \texttt{PerfGapAnalysis}(g_k)$}
    \STATE \highlightyellow{$\theta_{k+1} \gets \texttt{ParameterUpdate}(\theta_k, p_k)$}
\ENDFOR
\end{algorithmic}
\end{minipage}
\hrule

\end{figure*}

\subsection{Kernel-Level Optimization with MAIC-RL}\label{subsec:kernel-level}

\sys optimizes code through inference-time reinforcement learning with persistent memory. Each optimization problem is an environment interaction: the agent observes a program state, samples an optimization action, lowers it into executable code, receives scalar and textual feedback from validation and profiling, and updates a persistent \KB. Model weights do not change; the policy improves through memory retrieval and rewriting.

\subsubsection{Problem Formulation}

We formulate \sys as inference-time policy improvement over code-optimization tasks. At each rollout step, the actor proposes a transformation from the current state and retrieved memory, and the critic turns correctness and performance evidence into scalar and text-based rewards and updates the persistent \KB. \Cref{tab:maic-formulation} gives the notation.

\begin{figure*}[t]
\centering
\includegraphics[width=\textwidth]{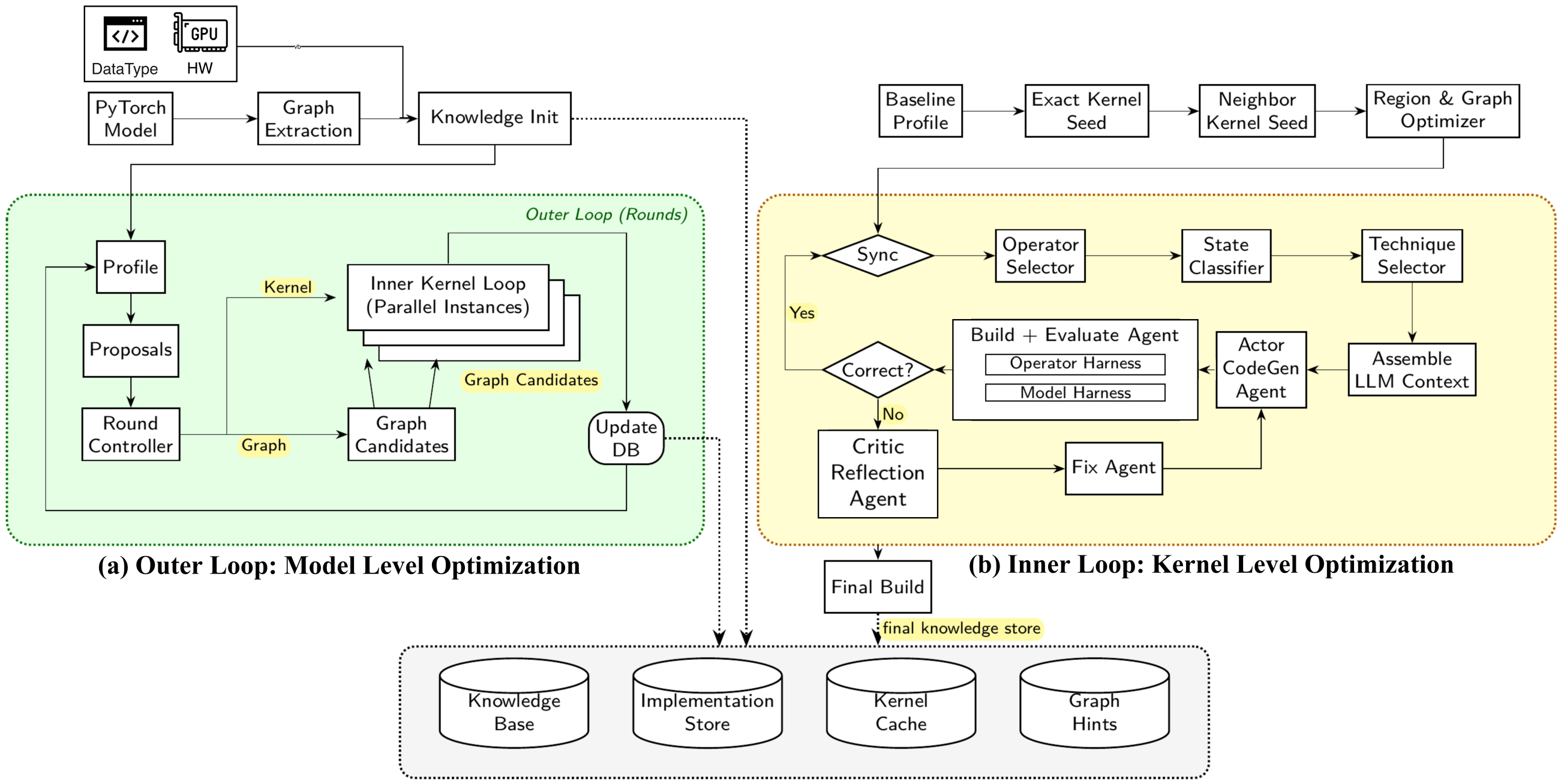}
\vspace{-20pt}
\caption{The two loops of \sys. (a) The outer loop extracts and profiles the model graph, proposes graph and kernel optimizations, and dispatches parallel inner loops that update shared databases. (b) The inner loop selects an operator, classifies its performance state, chooses a technique, and generates and validates code using operator and model harnesses. Failures trigger critic reflection and repair. Dotted arrows indicate writes to the shared knowledge base, implementation store, kernel cache, and graph hints.}

\label{fig:architecture}
\end{figure*}

\begin{table}[t]
\centering
\small
\caption{Problem formulation for \sys's Memory-Augmented In-Context Reinforcement Learning (MAIC-RL) loop.}
\label{tab:maic-formulation}
\renewcommand{\arraystretch}{1.3}
\begin{tabularx}{\linewidth}{@{}lX@{}}
\toprule
\textbf{Component} & \textbf{Definition} \\
\midrule
Task &
$\tau=(x,y,b,m)$, where $x$ is the input program or model fragment, $y$ is a reference implementation, $b$ is the backend, and $m$ is the measurement interface. \\

State &
$s_t=\phi(x_t,c_t,p_t,h_t)$, where $x_t$ is the current implementation, $c_t$ is compiler and validation feedback, $p_t$ is timing or profiling evidence, and $h_t$ is rollout history. \\

Actor action &
$a_t\sim\pi_\theta(a\mid s_t,\mem_t)$, where $a_t$ is a code transformation or implementation plan and $\mem_t$ is the set of retrieved \KB entries. \\

Reward signal &
$(r_t,z_t)=R(x_{t+1},y,b,m)$, where $r_t$ is the correctness-gated scalar performance reward and $z_t$ is the text-based reward signal. \\

Memory update &
$\mem_{k+1}=U\!\left(\mem_k,\tau,\{(s_t,a_t,r_t,z_t)\}_{t=0}^{T}\right)$, where the critic incorporates rollout outcomes into persistent long-term memory. \\
\bottomrule
\end{tabularx}
\end{table}

The scalar reward $r_t$ captures correctness-gated performance, and the text-based reward $z_t$ explains the performance gap, failure mode, violated assumptions, and target constraints. The policy parameters available to the LLM at inference time therefore include the retrieved and updated \KB as well as the prompt template.

\subsubsection{Hardware-Aware Profiling and Performance States}
\label{subsec:profiling}

After compilation and correctness validation, the harness profiles each candidate using backend-specific tools: Nsight Compute (NCU)~\cite{nsightcompute} on CUDA provides a 15-dimensional signature capturing throughput, occupancy, cache behavior, pipeline utilization, and warp scheduling; RVV exposes hardware counters and cycle counts; and Qualcomm GPUs use per-operator OpenCL~\cite{khronosOpenCL} event timing with fewer available signals. The State Classifier maps these signatures to hierarchical bottleneck states (e.g., memory-bound/cache thrashing or latency-bound/low parallelism), conditioned on hardware constraints. These states guide technique retrieval, while the critic uses profiling evidence to generate performance-gap critiques as text-based rewards. On KernelBench Level~2, NCU feedback improves speedup from 1.22$\times$ with latency-only feedback to 1.57$\times$ (Appendix~\ref{app:profiling_fidelity}). Clustering NCU signatures also identifies six interpretable optimization regimes (Appendix~\ref{sec:analysis}).

The Technique Selector combines shared techniques (coalescing, vectorization, tiling, unrolling, reductions, scans) with architecture-specific optimizations. NVIDIA techniques include \texttt{mma.sync} (Ampere/Ada), warpgroup MMA and TMA (Hopper), and \texttt{tcgen05} with Tensor Memory (Blackwell), informed by SM count, shared memory, L2 capacity, DRAM bandwidth, and tensor-core throughput. Adreno techniques include texture-path reads, register tiling, full-wave subgroups, and FP16 arithmetic (2$\times$ FP32 throughput). RVV techniques include vector-length-agnostic strip mining with LMUL, segment loads, widening multiply-accumulate, and vector reductions. Starting from only four RVV techniques, the knowledge base expands through measured outcomes, enabling optimization on targets with limited existing code (Section~\ref{sec:evaluation}).

\subsubsection{Persistent Long-Term Memory}

The \KB stores performance states, techniques, hypotheses, evidence-backed claims, outcomes, and verified implementations. Its compact optimization entries take the form
\[
e_i = \langle s_i, (a_i, q_i, z_i, c_i) \rangle,
\]
where $s_i$ denotes the performance state, $a_i$ the optimization action, $q_i$ its measured or predicted score, $z_i$ the textual critique, and $c_i$ the target, numerical, and failure constraints. Rather than relying on source-code similarity, retrieval matches performance states and hardware scopes, enabling reuse across different kernels and targets while respecting architecture-specific constraints. Scoped semantic updates preserve successful and failed transformations, their causes, and performance changes across optimization trajectories, providing compact, transferable experience without storing full programs for every attempt.

\subsubsection{Actor--Critic Textual Gradient Descent}

\sys instantiates MAIC-RL as \emph{Actor-Critic Textual Gradient Descent} (AC-TGD), an inference-time learning architecture in which natural-language critiques serve as policy-update signals. AC-TGD follows the REINFORCE-like ICRL procedure in Algorithm~\ref{alg:ICRL}. Instead of updating weights by gradient descent, it treats the LLM context and the retrieved \KB entries as mutable policy parameters $\theta$. Two kinds of update act on them: the scalar reward updates the statistics that rank techniques, and textual gradients edit the claims, states, and hypotheses that the actor reads. Figure~\ref{fig:architecture}(b) shows the inner loop that implements this flow.

\begin{table*}[t]
  \caption{Components of the \sys agentic workflow and their counterparts in the RL formulation.}
  \small
  \label{tab:workflow_components}
  \renewcommand{\arraystretch}{1.3}
  \begin{tabularx}{\textwidth}{@{}ll>{\raggedright\arraybackslash}X@{}}
  \toprule
  \textbf{Component} & \textbf{RL counterpart} & \textbf{Functionality} \\
  \midrule
  Knowledge Base & Parameters $\theta$ & Store of states, techniques, hypotheses, claims, outcomes, and verified implementations; with the prompt it forms the policy parameters. \\
  Operator Selector & & Picks the next kernel slot by time share, stall decay, and exploration. \\
  State Classifier & State $s_t$ & Assigns the deepest matching performance state from the profiling signature, with the program, feedback, and rollout history. \\
  Technique Selector & Policy $\pi_\theta(a_t \mid s_t)$ & Picks a technique for the state with a UCB rule over reward statistics. \\
  Actor CodeGen Agent & \multirow{2}{*}{Action $a_t$} & Rewrites the selected slot using claims, similar kernels, and past attempts. \\
  Fix Agent & & Repairs candidates that fail the correctness check, revising the action. \\
  Build + Evaluate Agent & Reward $r_t$ & Checks correctness and times against the incumbent; the gain is the scalar reward. \\
  Critic Reflection Agent & Policy evaluation & Every eight outcomes, checks proposed lessons against the records they cite; supported lessons are the text-based reward $z_t$. \\
  Knowledge update & Policy update & Logs each outcome, updates reward statistics, and applies validated edits to claims, states, and hypotheses. \\
  \bottomrule
  \end{tabularx}
\end{table*}

\textbf{Technique Selector.}
In each inner-loop step, the Operator Selector samples a kernel slot based on its partition-time contribution, stagnation decay, and exploration bonus. The State Classifier identifies its bottleneck, and the Technique Selector uses an upper confidence bound (UCB) multi-armed bandit to balance exploitation of effective techniques against exploration of uncertain ones:
\[
a_t=\arg\max_a\left[
\hat{\mu}_{s,a}
+c\sqrt{\frac{\ln(N_s+1)}{N_{s,a}+1}}
\right],
\]
where $\hat{\mu}_{s,a}$ is the estimated reward for technique $a$ in state $s$, $N_{s,a}$ its selection count, $N_s$ the total state visits, and $c$ the exploration coefficient. Sparse estimates shrink toward parent-scope statistics to transfer prior evidence across targets.

\textbf{Actor CodeGen Agent and Fix Agent.}
The actor rewrites only the selected slot using neighboring signatures, reference code, tensor shapes, stored implementations, retrieved claims, and search history. Smaller models handle mechanical optimizations and stronger models handle structural transformations. When a candidate fails, the Fix Agent revises it in a bounded repair loop and escalates the final attempt to the stronger model. Both produce the action $a_t$.

\textbf{Build + Evaluate Agent.}
Correct candidates are compared against the incumbent using alternating, same-process CUDA-graph replay of the deployed partition. The gain is $1-\operatorname{median}(T_{\mathrm{cand}}/T_{\mathrm{best}})$. Gains of at least 1\% are remeasured against the best composite of all slots, with $r_t$ defined as the smaller measured gain; smaller changes receive neutral reward.



\textbf{Critic Reflection Agent.}
After every eight outcomes, the critic reviews technique applications, state transitions, metric changes, speedups, and errors. It proposes evidence-backed edits covering technique effectiveness, pitfalls, interactions, parameters, state refinements, and performance explanations. Unsupported edits are rejected; accepted claims track supporting and contradicting evidence, decay over time, and retire when contradicted. Claims are scoped by hardware, operator, state, or technique and retrieved in subsequent prompts, forming the text-based reward $z_t$ alongside scalar reward $r_t$.

\subsubsection{Uniform Harness and Cross-Task Instantiation Across Targets}

To test whether the same persistent memory and text-based reward support cross-task reasoning, we instantiate the same actor and critic loop on problems whose targets differ in instruction-set architecture (ISA), source language, and execution model: CUDA C++ on NVIDIA GPUs, OpenCL C on Qualcomm GPUs, and RVV intrinsic code on RISC-V vector cores. All three share the MAIC-RL and AC-TGD learning schema, the \KB, the Actor CodeGen Agent's prompt schema, the Critic Reflection Agent's textual-gradient format, and the memory-retrieval mechanism. Only lowering, compilation, validation, and profiling are backend-specific, which isolates the learning contribution from each target's engineering. Appendix~\ref{app:code_samples} shows generated code for backend-specific transformations the agent discovers, such as Tensor Core utilization, MMA layout, and software warp specialization.
\subsection{From Kernels to Whole Models}
\label{subsec:arch}

Figure~\ref{fig:architecture} illustrates how the kernel agent integrates with the whole-model optimization flow. Graph Extraction converts a PyTorch model, weights, and reference I/O into a graph encoding operator dependencies, shapes, and layouts for target-specific code generation. Knowledge Init initializes the Knowledge Base using hardware and datatype information. Each outer-loop round profiles the current graph, generates graph and kernel proposals, and ranks them by measured per-operator latency. Kernel tasks and graph candidates are dispatched to parallel Inner Kernel Loops, while Update DB persists verified implementations and learned optimization knowledge for subsequent rounds.

Unlike fixed-template compilers, LLM-generated kernels have unknown costs until generated, validated, and profiled. Thus, \sys minimizes measured whole-model latency under a search budget spanning LLM inference, compilation, and hardware evaluation. The outer loop evaluates original and transformed variants back-to-back under identical seeds and kernel-search budgets, retaining changes only when they improve whole-model latency without violating accuracy constraints. Accepted changes take effect in the next round; optimization terminates when no changes are accepted and no new tasks remain, followed by end-to-end verification. Kernel specialization can reverse earlier graph decisions, while the search budget limits candidate exploration. Currently, \sys prioritizes partitions by measured cost; learned ranking remains future work.

\paragraph{Inner Kernel Loop.}
Each instance runs the kernel agent (Section~\ref{subsec:kernel-level}) on one task, initialized by \textit{Baseline Profile}, \textit{Exact Kernel Seed}, \textit{Neighbor Kernel Seed}, and \textit{Region \& Graph Optimizer}. Each iteration begins at \textit{Sync} with shared databases. \textit{Operator Selector} chooses a slot, \textit{State Classifier} identifies its profiling state (Section~\ref{subsec:profiling}), and \textit{Technique Selector} retrieves an optimization from the \KB. \textit{Assemble LLM Context} constructs the prompt for \textit{Actor CodeGen Agent}, whose output is compiled and validated by \textit{Build + Evaluate Agent} using the \textit{Operator Harness} (reference correctness) and \textit{Model Harness} (whole-model correctness). Valid candidates return to \textit{Sync}; failures invoke \textit{Critic Reflection Agent} and \textit{Fix Agent} for repair and regeneration. Finally, \textit{Final Build} emits the selected kernel, and the final knowledge store persists new knowledge.

\paragraph{Graph Hints.}
All graph changes pass through a versioned \texttt{graph\_hints/v1} schema in the \textit{Graph Hints} database. Each hint specifies a transformation (e.g., fusion, splitting, or core-width assignment), affected operators, and its source (\textit{Inner Kernel Loop} or \textit{Proposals}). A shared backend-aware routine applies each rewrite, validates reference correctness, and measures performance. Independent hints are evaluated in parallel, while cached acceptance and rejection decisions support resumable execution.

\subsection{Modular Interfaces}
\label{subsec:modular}

The \textit{Outer Loop} and \textit{Inner Loop} run as separate processes, exchanging on-disk \textit{Graph Hints}, \textit{Kernel Cache}, \textit{Implementation Store}, \textit{Knowledge Base}, per-dispatch profiles, and kernel-task plans. Either loop operates independently: the inner loop optimizes standalone kernels, while the outer loop can use a fixed kernel library. An external graph compiler can replace the graph optimizer by consuming \textit{Graph Hints} and returning graph mappings and measurements. New frameworks and hardware targets require only a frontend or declarative backend description, isolating target-specific behavior.

\section{Evaluation}\label{sec:evaluation}
We organize our cross-target and baseline comparisons around four research questions: 
\textbf{Q1.} Are the resulting implementations competitive with existing baselines?
\textbf{Q2.} Does persistent memory improve cross-task optimization? 
\textbf{Q3.} Do text-based reward signals carry information beyond a scalar reward?
\textbf{Q4.} Does the same memory schema transfer across heterogeneous targets?

\begin{table*}[t]
\centering
\caption{\sys's speedup across NVIDIA, Qualcomm, and RISC-V
backends. Baselines are PyTorch eager~\cite{ansel2024pytorch2} (NVIDIA), QNN SDK
(Adreno), and scalar kernels or ExecuTorch/XNNPACK (RISC-V).
Share denotes the percentage of solved benchmark problems.}
\label{tab:summary_all}
\fontsize{7.5}{9}\selectfont
\setlength{\tabcolsep}{3.5pt}
\renewcommand{\arraystretch}{1.12}

\begin{tabular}{@{}p{2.65cm} c l c c c c c c c c@{}}
\toprule
Hardware / Benchmark & Level & Method & Share & Mean & GeoM &
Med & Min & Max & \%$>$1x & \%$<$1x \\
\midrule

\multirow{7}{*}{\parbox{2.65cm}{\raggedright
\textit{NVIDIA L40S}\\
\textit{KernelBench}}}
& \multirow{3}{*}{1}
& AI CUDA Engineer & 82\% & 1.728 & 1.101 & 1.202 & 0.0659 & 32.35 & 74.39\% & 25.61\% \\
& & StitchCUDA & 98\% & 1.718 & 1.335 & 1.116 & 0.0250 & 9.77 & 86.73\% & 11.22\% \\
& & \textbf{Ours} & 96\% & 3.972 & 1.850 & 1.758 & 0.2621 & 158.51 & 85.42\% & 14.58\% \\
\cmidrule(lr){2-11}

& \multirow{3}{*}{2}
& AI CUDA Engineer & 83\% & 3.865 & 1.695 & 1.574 & 0.328 & 123.39 & 79.52\% & 20.48\% \\
& & StitchCUDA & 79\% & 2.888 & 2.177 & 1.665 & 0.296 & 9.48 & 94.94\% & 5.06\% \\
& & \textbf{Ours} & 100\% & 22.823 & 5.576 & 4.934 & 0.8084 & 314.64 & 85.00\% & 15.00\% \\
\cmidrule(lr){2-11}

& 3 & \textbf{Ours} & 100\% & 8.492 & 5.182 & 4.571 & 0.9016 & 30.42 & 97.83\% & 2.17\% \\

\midrule

\multirow{2}{*}{\parbox{2.65cm}{\raggedright
\textit{NVIDIA B200}\\
\textit{SOL-ExecBench}}}
& 1 & \textbf{Ours} & 71\% & 4.486 & 2.241 & 2.080 & 0.115 & 33.64 & 76.12\% & 23.88\% \\
& 2 & \textbf{Ours} & 59\% & 2.925 & 1.138 & 1.013 & 0.166 & 45.20 & 52.08\% & 47.92\% \\

\midrule

\parbox{2.65cm}{\raggedright
\textit{Qualcomm Adreno 650}\\
\textit{KernelBench}}
& 1 & \textbf{Ours} & 43\% & 17.57 & 5.800 & 9.180 & 0.0142 & 73.77 & 85.00\% & 15.00\% \\

\midrule

\multirow{2}{*}{\parbox{2.85cm}{\raggedright
\textit{RISC-V Saturn V256D128}\\
\textit{KernelBench}}}
& \multirow{2}{*}{1}
& \textbf{Ours}, vs scalar (83) & 83\% & 16.533 & 6.989 & 6.786 & 1.000 & 111.19 & 97.59\% & 0.00\% \\
& & \textbf{Ours}, vs XNNPACK (70) & 70\% & 70.410 & 2.740 & 1.394 & 0.0189 & 934.19 & 62.86\% & 37.14\% \\

\bottomrule
\end{tabular}
\end{table*}

\subsection{Experimental Setup}

We evaluate \sys on KernelBench Levels~1--3~\cite{ouyang2024kernelbench} and SOL-ExecBench~\cite{lin2026solexecbench} across CUDA/NVIDIA, OpenCL~\cite{khronosOpenCL}/Qualcomm, and RVV/RISC-V, using a shared actor--critic loop and memory schema with target-specific compilation, validation, and profiling. SOL-ExecBench contains 235 problems from 124 AI models targeting Blackwell; we evaluate Level~1 (94 single-operation blocks) and Level~2 (82 fused blocks) in BF16/FP32, reporting speedup over PyTorch eager rather than speed-of-light scores.

Valid solutions must compile, execute, and pass reference correctness checks; performance statistics include only valid solutions. We report $\mathit{fast}_p$ as the fraction of correct solutions achieving at least $p$ speedup, unlike KernelBench's denominator over all problems. NVIDIA results compare against PyTorch eager and prior CUDA agents; Qualcomm results against QNN SDK and Apache TVM~\cite{chen2018tvm} using on-device GPU timing; and RVV results against ExecuTorch/XNNPACK and scalar kernels, measured in cycles on a single-core Saturn V256D128~\cite{zhao2024instruction} simulated in FireSim~\cite{karandikar2018firesim}. Table~\ref{tab:summary_all} summarizes results, and a soft-verification agent prevents PyTorch dispatch and external-library shortcuts (Appendix~\ref{app:correctness}).

\begin{figure*}[!h]
    \centering
        \includegraphics[width=0.9\linewidth]{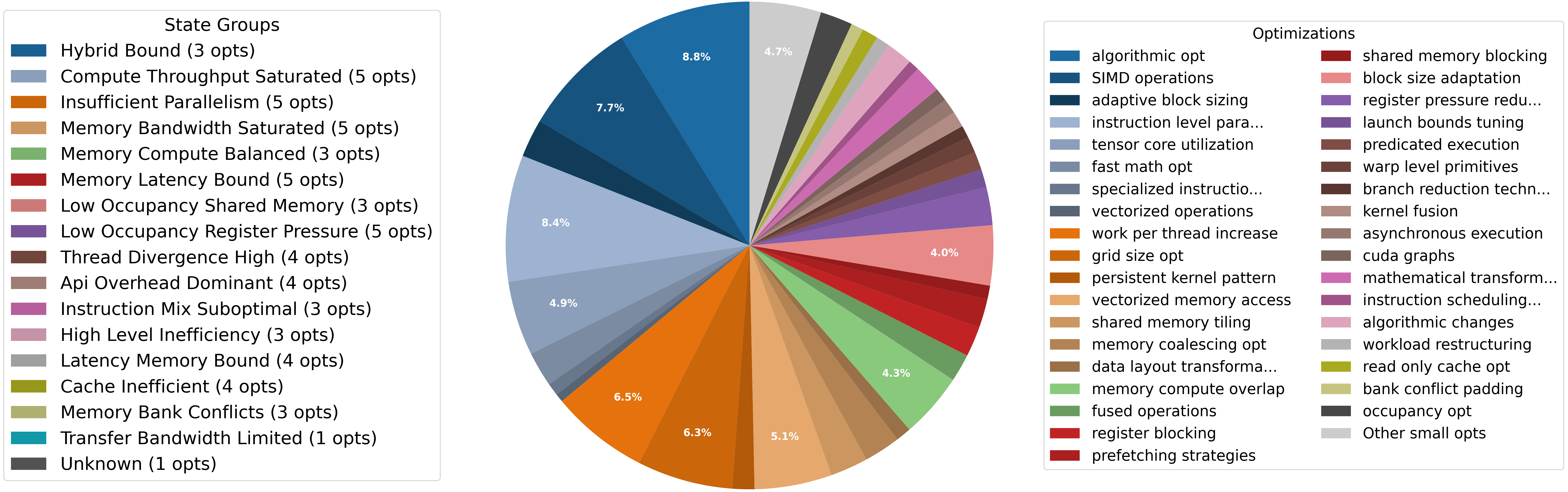}
        \vspace{-10pt}
        \caption{Distribution of 3972 CUDA optimization applications across performance states.}
        \label{fig:opt_pie}

\end{figure*}

\newcolumntype{P}[1]{>{\raggedright\arraybackslash}p{#1}}
\begin{table*}[t]
\centering
\caption{Comparison among \sys (Ours), AI CUDA Engineer, and StitchCUDA.}
\label{tab:cuda_vs_tachyon_iree}
\scriptsize
\renewcommand{\arraystretch}{1.35}
\resizebox{\linewidth}{!}{%
\begin{tabular}{@{}P{1.3cm}P{1cm}P{2cm}P{2cm}P{2.4cm}P{2.1cm}P{5.1cm}@{}}
\toprule
{System} &
{Language} &
{Agent's Models} &
{Hardware Setup} &
{Dataset} &
{Hyperparameter} &
{Test Harness} \\
\midrule
\rowcolor{RowBlue}
\sys (Ours) &
CUDA &
GPT-5.0-mini, GPT-4.1, GPT-5.0, Sonnet 4.6, Opus 4.6 &
NVIDIA A6000, A100, H100, L40S; RISC-V Vector core (Saturn) &
KernelBench L1--3, SOL-ExecBench &
10 iterations, 10 rollout steps per iteration &
\cellcolor{CellYellow}\textbf{Functionality Check}: passing kernels are compiled and executed on GPUs; outputs are compared to PyTorch baselines with multiple randomized seeds to ensure correctness and prevent overfitting. \\

\rowcolor{RowGreen}
AI CUDA Engineer &
CUDA &
O4-mini, Sonnet 3.7, Gemini 2.5 Pro, GPT-4.1, O3 &
NVIDIA H100 &
KernelBench L1--2 &
10 generations; 10 proposals/generation; top 4 evaluated &
\cellcolor{CellYellow}\textbf{LLM Soft-Verification}: an LLM scans each kernel for likely compilation, memory, or numerical errors, filtering bad code before GPU execution. \\

\rowcolor{RowOrange}
StitchCUDA &
CUDA &
Opus 4.6 (our run) &
NVIDIA L40S (our run) &
KernelBench L1--2 &
15 attempts per problem with 2 replans &
\cellcolor{CellYellow}\textbf{Verifier agent}: checks correctness and profiles with Nsys and NCU; speedup over PyTorch is measured as wall-clock time in its own harness. \\

\rowcolor{RowGray}
QNN SDK Compiler &
OpenCL &
N/A &
Qualcomm Adreno 650 &
KernelBench L1 &
N/A &
\cellcolor{CellYellow}\textit{(SDK-internal validation)} \\

\bottomrule
\end{tabular}%
}
\end{table*}

\subsection{Evaluating Generality across Hardware Generations}

Level~2 gains are larger than Level~1 gains on the L40S because composed operators allow multi-step transformations such as fusion and layout changes. Level~1 problems are single operators with nothing to fuse, and the KernelBench harness does not allow ahead-of-time weight packing.

\begin{figure}[]
\vspace{-10pt}
    \centering
    \includegraphics[width=\columnwidth]{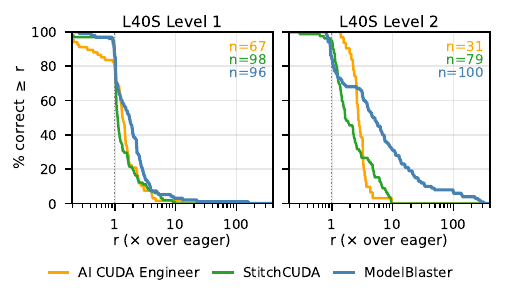}
    \vspace{-30pt}
    \caption{$\mathit{fast}_p$ curves on an L40S for KernelBench Levels~1 and 2: the share of problems with a correct result whose speedup over PyTorch eager is at least $r$. All three systems are measured as end-to-end wall-clock time. $n$ is the number of problems with valid results; AI CUDA Engineer has kernels that directly call external libraries excluded. }
    \label{fig:fastp_pair}
    \vspace{-20pt}
\end{figure}

We ran StitchCUDA~\cite{li2026stitchcuda}, a multi-agent CUDA framework, with the same model (Opus 4.6), GPU, and problem sizes as ours and its search budget exhausted (15 attempts per problem). Over PyTorch eager, \sys reaches 1.85x on Level~1 and 5.58x on Level~2, against 1.335x and 2.177x for StitchCUDA~\cite{li2026stitchcuda} (Table~\ref{tab:summary_all} and Figure~\ref{fig:fastp_pair}). On Level~1, 33 of its 98 solutions return PyTorch operations or call libraries instead of generating kernels, and its 65 generated kernels average 1.496x. \sys gains more on Level~2 because its memory reuses fusion and layout transformations across kernels. We compare mainly on CUDA because most public agentic baselines target it, and the other targets test retargetability.

On the 115 SOL-ExecBench problems with a measured eager time, \sys reaches a 1.69$\times$ geometric-mean speedup over PyTorch eager (median 1.47$\times$). Gains concentrate on small problems: 4.03$\times$ on the small set, 1.30$\times$ on the large set, and 0.98$\times$ above our size cap.

\subsection{Evaluate Cross-Target Transferability}
We evaluate \sys beyond CUDA by running the same learning loop on OpenCL/Qualcomm and RVV/RISC-V which is programmed with different language, execution model, and measurement interface. 

On 42 KernelBench Level~1 problems on a Qualcomm Adreno~650, \sys reaches a 5.80x geometric-mean speedup over QNN (median 9.18x). The largest gains are on reduction-axis operators, where the learned strategy maps a reduction bottleneck to cooperative local-memory reduction and vectorized global I/O. As an optimization-based baseline we use Apache TVM~\cite{chen2018tvm} with its default Adreno pipeline (rule-based schedules with texture memory, no per-operator autotuning) on the same device. On the 39 operators that both systems cover, \sys is 1.69x faster in geometric mean and 2.30x faster on the 16 compute-bound operators (Appendix~\ref{app:tvm}). TVM is faster on BatchNorm, LayerNorm, and matrix-vector multiplication. Autotuning would improve TVM at a large offline search cost, which we leave to future work.

On RVV we measure cycles on a single-core Saturn V256D128~\cite{zhao2024instruction} in FireSim~\cite{karandikar2018firesim}. Against the scalar reference kernels, \sys reaches a 6.99x geometric-mean speedup over 83 problems (median 6.8x). Against ExecuTorch with XNNPACK, a production framework with hand-tuned RVV microkernels, it reaches 2.74x on the 70 problems that both cover (median 1.39x), and Appendix~\ref{app:rvv} lists the operator classes. \sys is faster on every class except convolution, where XNNPACK ships im2col and microkernel paths that the KernelBench harness does not let a generated kernel use. XNNPACK also has no kernel for 10 problems that \sys handles. On Level~1, \sys is therefore stronger on Qualcomm and RVV than on the L40S, where the harness limits above apply, and the same memory-augmented agent remains useful when the implementation language and hardware execution model change.

\subsection{Adapt to Full-Model and Subgraph Optimization}
On 46 KernelBench Level~3 models on an L40S, \sys reaches a 5.18$\times$ geometric-mean speedup over PyTorch eager (Table~\ref{tab:summary_all}). Gains are largest on LeNet5 (10.6$\times$), ResNetBasicBlock (3.2$\times$), and DenseNet121DenseBlock (2.7$\times$), and only about 1.1$\times$ on the plain MLPs. Level~1 and Level~2 memory entries carry over to Level~3 through cross-layer fusion, activation reuse, and algebraic simplification. Per-model results and generated code are in Appendix~\ref{sec:appendix_level3}. Whole-model code generation adds long boilerplate, heterogeneous bottlenecks, and weaker performance signals. Hierarchical decomposition, where the critic selects subgraphs or kernels for further optimization, addresses this, and the model-level optimizer (Section~\ref{sec:method}) takes this role.

\subsection{Distribution of Optimizations learned in Persistent Memory }\label{sec:opt-dist}

We analyze the \KB's composition and cross-task reuse potential. The active learned \KB occupies approximately 10 MB and organizes techniques by performance state. Across 3,972 optimization applications on KernelBench L1+L2 (A6000), no state accounts for more than 20\% of applications (Figure~\ref{fig:opt_pie}), and kernels traverse 5.5 states on average, requiring state reassessment throughout optimization.

L40S trajectories reveal diminishing returns from repeated micro-tuning (e.g., \texttt{instruction\_level\_parallelism}, \texttt{grid\_size\_optimization}), but stronger gains from ordered cross-family transformations: \texttt{shared\_memory\_tiling} before \texttt{tensor\_core\_utilization} achieves 2.41$\times$ median speedup, and layout transformation before fusion achieves 1.95$\times$. The \KB preserves these action sequences and state transitions, which cannot be derived directly from the final implementation.

\section{Conclusion}

We present \sys, a memory-augmented in-context reinforcement learning framework for code optimization across heterogeneous accelerators. Rather than updating model weights, \sys uses execution-grounded textual feedback to maintain a persistent knowledge base of optimization states, actions, outcomes, and constraints. Its actor--critic loop accumulates reusable experience across tasks, while a model-level optimizer extends kernel optimization to graph transformations guided by measured whole-model performance.

Across NVIDIA GPUs, Qualcomm Adreno GPUs, and RISC-V vector cores, \sys achieves substantial speedups over existing frameworks and optimization agents without target-specific training data. These results demonstrate the effectiveness of a shared optimization methodology across programming languages, architectures, and profiling interfaces, including low-resource targets. By connecting kernel-level learning with graph-level decisions, \sys provides a unified approach to optimization beyond individual kernels.



\bibliographystyle{mlsys2025}
\bibliography{references}

\clearpage
\onecolumn
\appendix
\label{sec:appendix}

\section*{Appendix}
This appendix is organized as follows. Appendix~\ref{app:baselines} gives additional baseline comparisons: TVM and QNN on the Adreno 650, and ExecuTorch with XNNPACK on RVV. Appendix~\ref{app:ablations} reports ablations of the memory, the profiling data, the agent design, and the flow components. Appendix~\ref{sec:analysis} contains additional analyses of search breadth, cost, optimization distributions, state clustering, verification, and limitations. Appendix~\ref{app:code_samples} shows generated code and per-model results.

\section{Additional Baseline Comparisons}\label{app:baselines}

\subsection{Adreno 650: TVM and QNN}\label{app:tvm}

On the Qualcomm Adreno 650, \sys is compared with QNN, the commercial OpenCL toolchain of Qualcomm, on 52 KernelBench Level~1 operators in FP16. We add Apache TVM (\texttt{opencl -device=adreno}) as the open-source, optimization-based baseline that emits OpenCL for the same device. TVM uses its default Adreno pipeline, with rule-based schedules and texture memory and without per-operator autotuning. It covers 42 of the 52 operators, 34 on the optimized pipeline and 8 through fallback schedules. Table~\ref{tab:tvm_qnn} reports pairwise geometric-mean speedups on the operators that both systems cover.
\vspace{-20pt}
\begin{table}[h]
\centering
\small
\caption{Pairwise geometric-mean speedups on the Adreno 650 on KernelBench Level~1 (FP16). $n$ is the number of operators covered by both systems. Compute-bound operators are a subset of the overlap.}
\label{tab:tvm_qnn}
\begin{tabular}{@{}lcc@{}}
\toprule
Comparison & All overlapping & Compute-bound \\
\midrule
\sys over TVM & 1.69$\times$ ($n=39$) & 2.30$\times$ ($n=16$) \\
\sys over TVM, optimized pipeline only & 1.50$\times$ ($n=32$) & \\
\sys over QNN & 5.80$\times$ ($n=42$) & \\
TVM over QNN & 3.36$\times$ ($n=37$) & 1.21$\times$ ($n=8$) \\
\bottomrule
\end{tabular}
\end{table}
\vspace{-20pt}

TVM is faster than \sys on BatchNorm, LayerNorm, and matrix-vector multiplication, where the gap reaches 0.3$\times$ for the matrix-vector case. These losses are included in the geometric means.

\subsection{RVV: ExecuTorch with XNNPACK}\label{app:rvv}

\begin{figure}[h]
\centering
\includegraphics[width=0.5\linewidth]{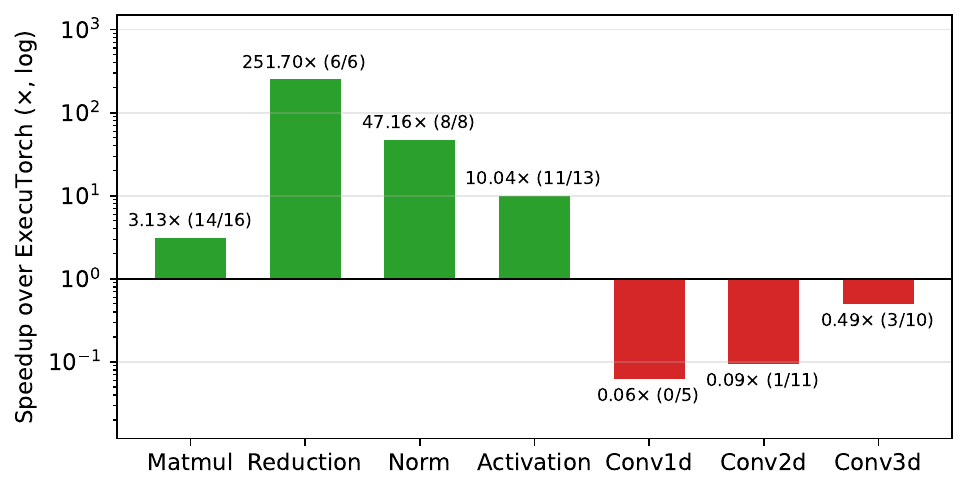}
\caption{Geometric-mean speedup of \sys over ExecuTorch with XNNPACK on RVV by operator class (log scale; values above 1 mean \sys is faster). Labels give the speedup and the number of problems where \sys is faster out of the class total.}
\label{fig:rvv_classes}
\end{figure}

On RISC-V vector cores we measure cycles on a single-core Saturn V256D128 in FireSim and compare with ExecuTorch with XNNPACK, a production framework with hand-tuned RVV microkernels. On the 70 problems that both systems cover, \sys is faster on 44 (63\%) with a 2.74$\times$ geometric-mean speedup and a median of 1.39$\times$. Table~\ref{tab:rvv_classes} and Figure~\ref{fig:rvv_classes} break the comparison down by operator class.

\begin{table}[h]
\centering
\small
\caption{\sys against ExecuTorch with XNNPACK on RVV by operator class: the number of problems where \sys is faster, and the geometric-mean speedup over ExecuTorch.}
\label{tab:rvv_classes}
\begin{tabular}{@{}lcc@{}}
\toprule
Operator class & \sys faster & Speedup \\
\midrule
Matrix multiplication & 14 of 16 & 3.13$\times$ \\
Reductions & 6 of 6 & 251.7$\times$ \\
Normalizations & 8 of 8 & 47.2$\times$ \\
Activations & 11 of 13 & 10.0$\times$ \\
Conv2d & 1 of 11 & 0.09$\times$ \\
Conv1d & 0 of 5 & 0.06$\times$ \\
Conv3d & split by shape class & 0.49$\times$ \\
\bottomrule
\end{tabular}
\end{table}

The advantage on activations arises because this XNNPACK build has no tuned activation kernels. The losses on convolutions come from the KernelBench harness. It allows no ahead-of-time weight packing, no per-shape microkernel dispatch, and no scratch tensor, while XNNPACK ships im2col and microkernel paths tuned for these shapes.

\section{Additional Ablation Studies}\label{app:ablations}
\subsection{Impact of Learning on Performance}\label{app:learning_rate}
The memory, profiling, and minimal-agent ablations in this section were run with an earlier version of the flow on KernelBench Level~1 and 2. Their numbers are not comparable with the L40S sweeps in Section~\ref{sec:evaluation}. The last subsection of this section reports the current flow.
To understand the impact of {\sys}' \KB on code optimization performance, we conduct a study by limiting the ability of  {\sys}' MAIC-RL flow to learn from experience effectively. As described in Section~\ref{sec:opt-dist}, it is critical to develop and explore a wide range of optimization strategies when optimizing a diverse codebase. However, this diversity relies on {\sys}'s ability to mutate its \KB by modifying and expanding entries.

\begin{figure}[h]
    \centering
    \begin{subfigure}[t]{0.5\linewidth}
        \centering
        \includegraphics[width=\linewidth]{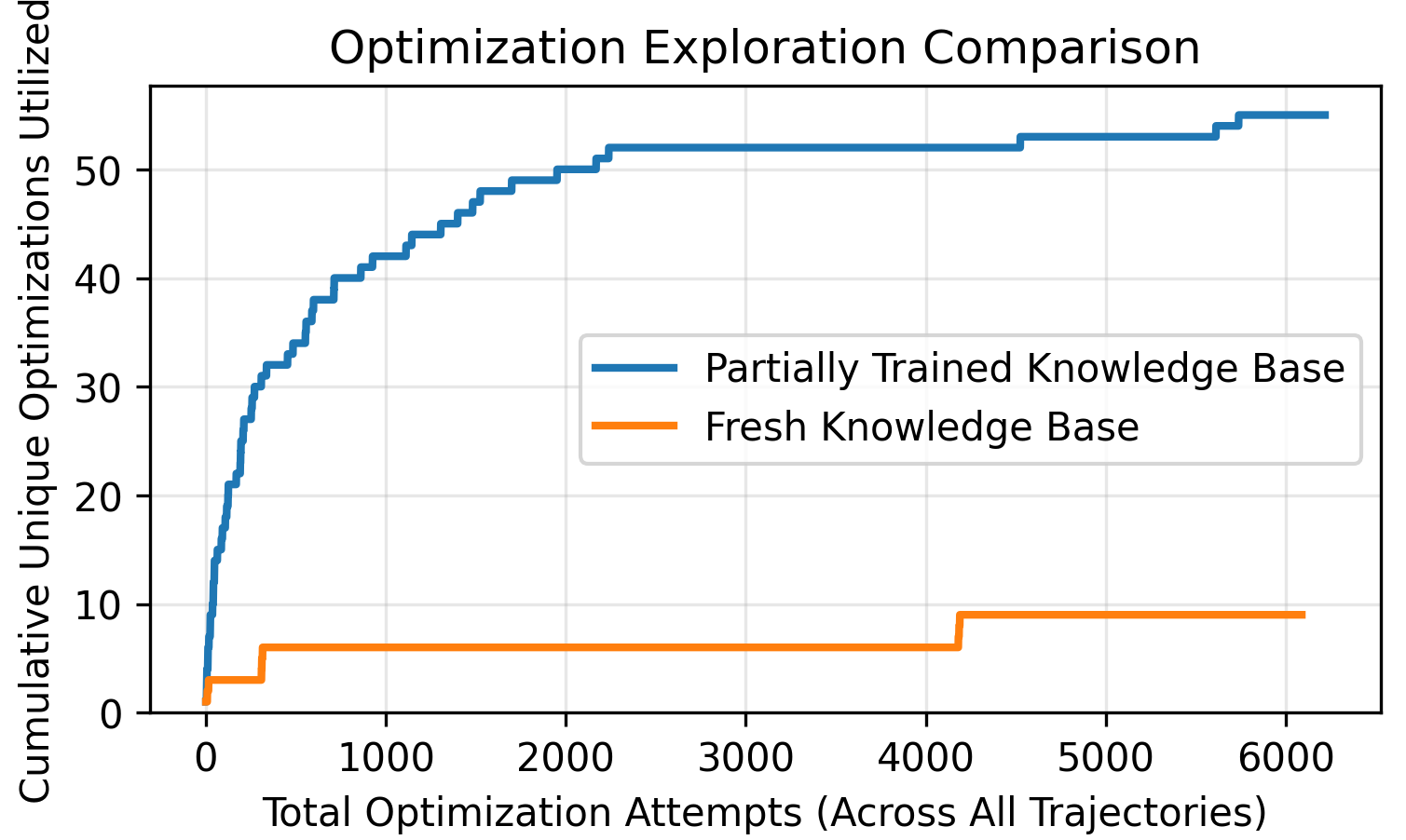}
        \caption{Pretrained vs.\ empty \KB.}
        \label{fig:learning_rate}
    \end{subfigure}
    \hfill
    \begin{subfigure}[t]{0.49\linewidth}
        \centering
        \includegraphics[width=\linewidth]{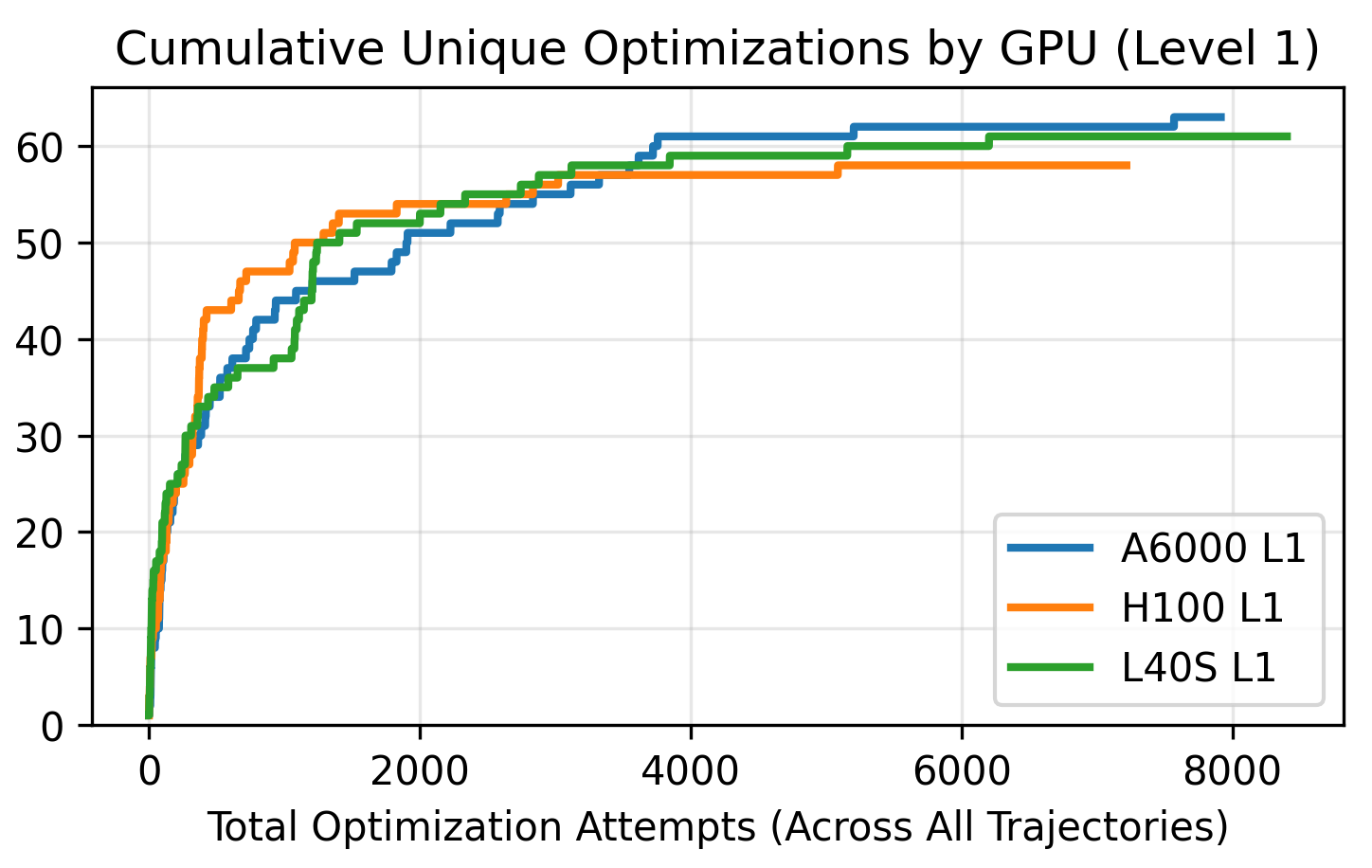}
        \caption{Pretrained \KB reused across GPUs.}
        \label{fig:knowledge_base-learning_rate}
    \end{subfigure}
    \caption{Reusable persistent memory: (a) discovery and application of new optimizations with a pretrained vs.\ empty \KB; (b) the same \KB trained on an A6000 transfers across other NVIDIA GPU platforms.}
    \label{fig:learning_rate_pair}
\end{figure}

We provide an example of states within the \KB in Figure~\ref{fig:kb-format}.
First, we study the impact of developing a \KB from scratch, compared to iterating upon a partially-trained data structure. The comparison between Figure~\ref{fig:learning_rate} and Figure ~\ref{fig:knowledge_base-learning_rate} shows the rate at which optimizations are applied while optimizing the KernelBench Level 1 dataset. As seen in the left plot, although the first round to construct a \KB is expensive, future optimization passes can benefit from much faster coverage of optimizations. Furthermore, these generated databases can be reused across scenarios;  The right plot demonstrates how a \KB trained on an A6000 GPU can be used for optimization runs across different GPU platforms.

To isolate the contribution of long-term memory from that of profiling, we further compare against a fresh-memory agent that has full target-specific profiling but starts with an empty \KB and no state-conditioned reuse. This agent is 1.67$\times$ slower than the full system. Profiling information alone therefore gives meaningful gains, and the strongest gains come from combining profile-grounded feedback with persistent, state-aware memory.

\subsection{Impact of Semantic Profiling Data on Agent Performance }\label{app:profiling_fidelity}
\begin{figure*}[t]
    \centering
    \begin{subfigure}[t]{0.4\textwidth}
        \centering
        \includegraphics[width=\columnwidth]{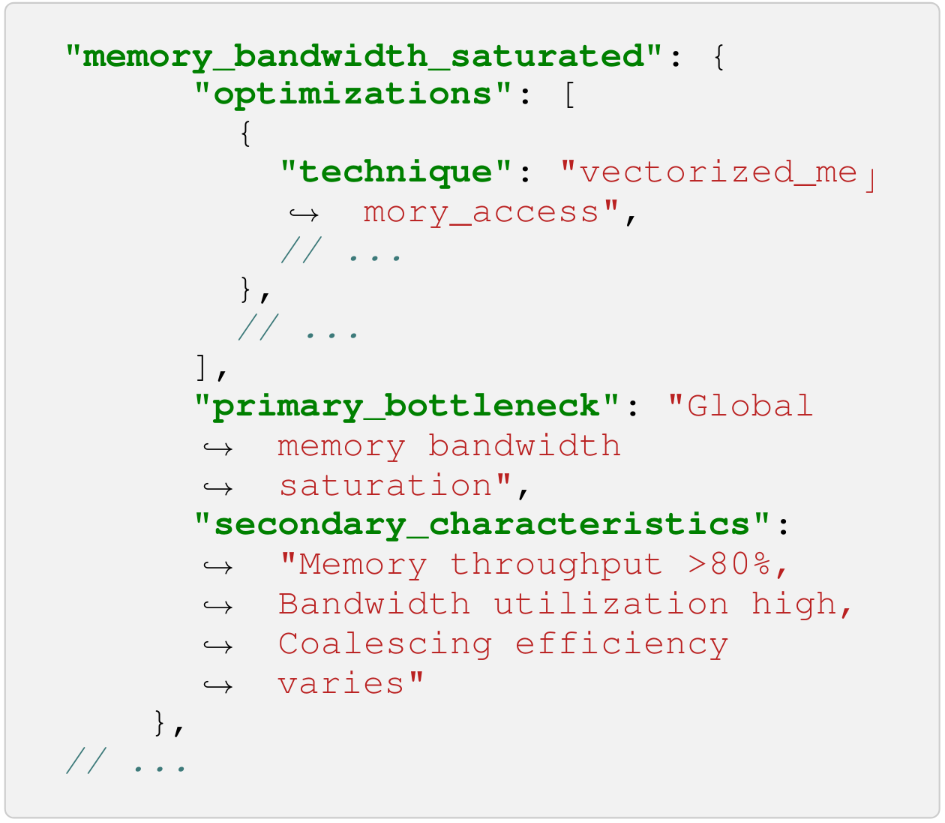}
        \caption{A sample of the \KB data structure, specifying discovered states.}
        \label{fig:kb-format}
    \end{subfigure}
    \hfill
    \begin{subfigure}[t]{0.59\textwidth}
        \centering
        \includegraphics[width=\linewidth]{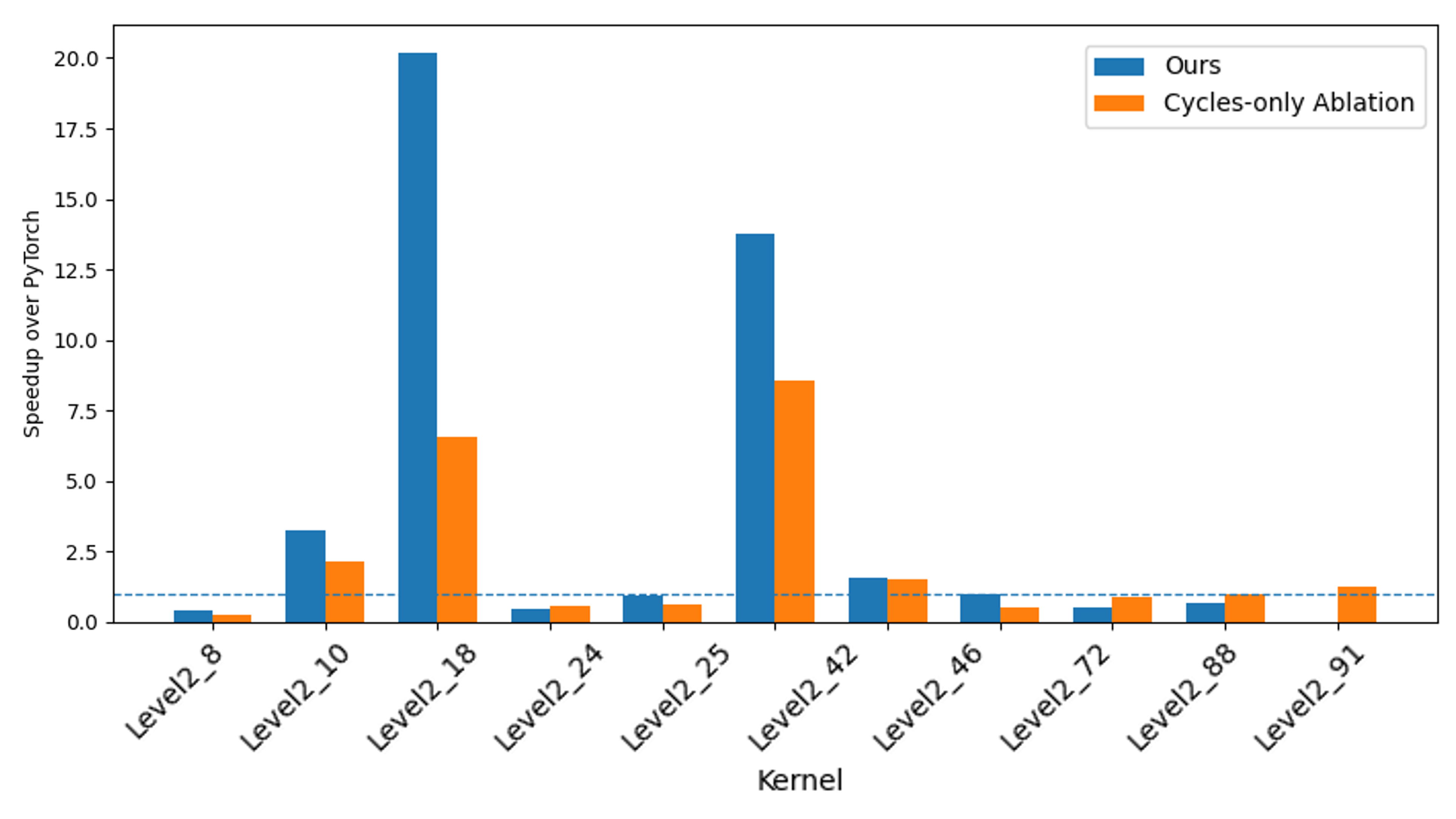}
        \caption{Speedup over PyTorch: full text-based reward vs.\ a cycles-only ablation.}
        \label{fig:ours_vs_ablation_cycles_only}
    \end{subfigure}
    \caption{Memory contents and text-based reward ablation. (a) The learned memory does not collapse to a single transformation family: no state exceeds 20\% of applications, justifying state-conditioned retrieval. (b) Replacing text-based critiques with cycle-only scalar feedback degrades optimization across most kernels (1.22$\times$ vs.\ 1.57$\times$ on Level~2), confirming that language carries information scalar timing cannot.}
    \label{fig:memory_textreward_pair}
\end{figure*}

To understand the impact of profiling fidelity, we conducted ablation studies against an Actor CodeGen Agent agent that only recieves latency rewards instead of detailed NCU profiles. The cycle-only agent underperforms approaches that use NCU (1.22x speedup over PyTorch on Level-2 compared to 1.57x with NCU data, shown in Figure~\ref{fig:ours_vs_ablation_cycles_only}), indicating that scalar latency alone is insufficient to infer why a kernel is slow or which optimization direction to pursue. Furthermore, in addition to limiting performance in the inner Actor CodeGen Agent loop, stripping semantic data prevents the Critic Reflection Agent from effectively learning from past experience.

\subsection{Comparison Against Minimal Agent}\label{app:minimal_agent}
Finally, we compare against a minimal agent. At each iteration, this agent directly takes in CUDA code and NCU profiling data and outputs optimized code. When running 10 trajectories of length 10, this agent requires 2.4x tokens compared to \sys. We observe this is due to two primary causes. First, because it has no access to a \KB and a guided reasoning process, this agent must devote more tokens up-front for reasoning about the optimizations. Second, the minimal agent requires more retrievals for correctness compared to \sys.

Compared to \sys, this minimal agent has 0.379x performance improvement per token. Furthermore, \sys achieves better performance in 71\% of cases compared to the simplified agent.

\subsection{Flow Components on Level 3 (L40S)}\label{app:flow_components}
We ablate the components of the model-level flow on eight Level~3 problems (VanillaRNNHidden, SqueezeNetFireModule, DenseNet121DenseBlock, DenseNet121, NetVladWithGhostClusters, NetVladNoGhostClusters, ResNetBasicBlock, and GoogleNetInceptionV1) on an L40S, with one seed per cell. A0 is the baseline flow. A1 adds the round controller, A2 adds a graph-first round (three rounds in total), A3 merges sibling convolutions, A4 adds region windows, and A5 adds the region loop. A6 combines the round controller, sibling-convolution merging, region windows, and the region loop. Table~\ref{tab:flow_arms} reports the geometric-mean speedup of the final model over PyTorch eager and over a CUDA graph. A6 is faster than eager on all eight problems and at least 2$\times$ faster on seven. With eight problems and one seed per cell, differences of about 10\% between arms are within the noise, and A0 and A1 gave the same geometric mean.

\begin{table}[h]
\centering
\small
\caption{Flow-component ablation on eight KernelBench Level~3 problems on an L40S: geometric-mean speedup of the final model over PyTorch eager and over a CUDA graph.}
\label{tab:flow_arms}
\begin{tabular}{@{}lcc@{}}
\toprule
Arm & vs eager & vs CUDA graph \\
\midrule
A0 baseline & 2.79$\times$ & 1.59$\times$ \\
A1 round controller & 2.79$\times$ & 1.59$\times$ \\
A2 graph-first round & 3.46$\times$ & 1.97$\times$ \\
A3 sibling-conv merge & 3.47$\times$ & 1.98$\times$ \\
A4 region windows & 4.08$\times$ & 2.33$\times$ \\
A5 region loop & 2.90$\times$ & 1.65$\times$ \\
A6 all of the above & 5.68$\times$ & 3.23$\times$ \\
\bottomrule
\end{tabular}
\end{table}

\section{Additional Analysis}\label{sec:analysis}

\subsection{Search Breadth and Search Depth}
The key search hyperparameters are trajectory count and trajectory length. Increasing trajectory count improves exploration, but we observe diminishing returns beyond eight trajectories for median and upper-quartile kernels. Lower-quartile kernels continue to benefit from additional breadth, suggesting that difficult kernels require broader exploration. Increasing trajectory length shows similar diminishing returns beyond four transformations for most kernels, although high-upside kernels can still benefit from longer sequences.

These results support the actor--critic structure. The actor needs enough breadth to explore transformations, while the critic should decide when additional rollouts are unlikely to pay off. This prevents the method from becoming an unbounded 

\begin{figure}[h]
    \centering
    \begin{subfigure}[t]{0.47\linewidth}
        \centering
        \includegraphics[width=\linewidth]{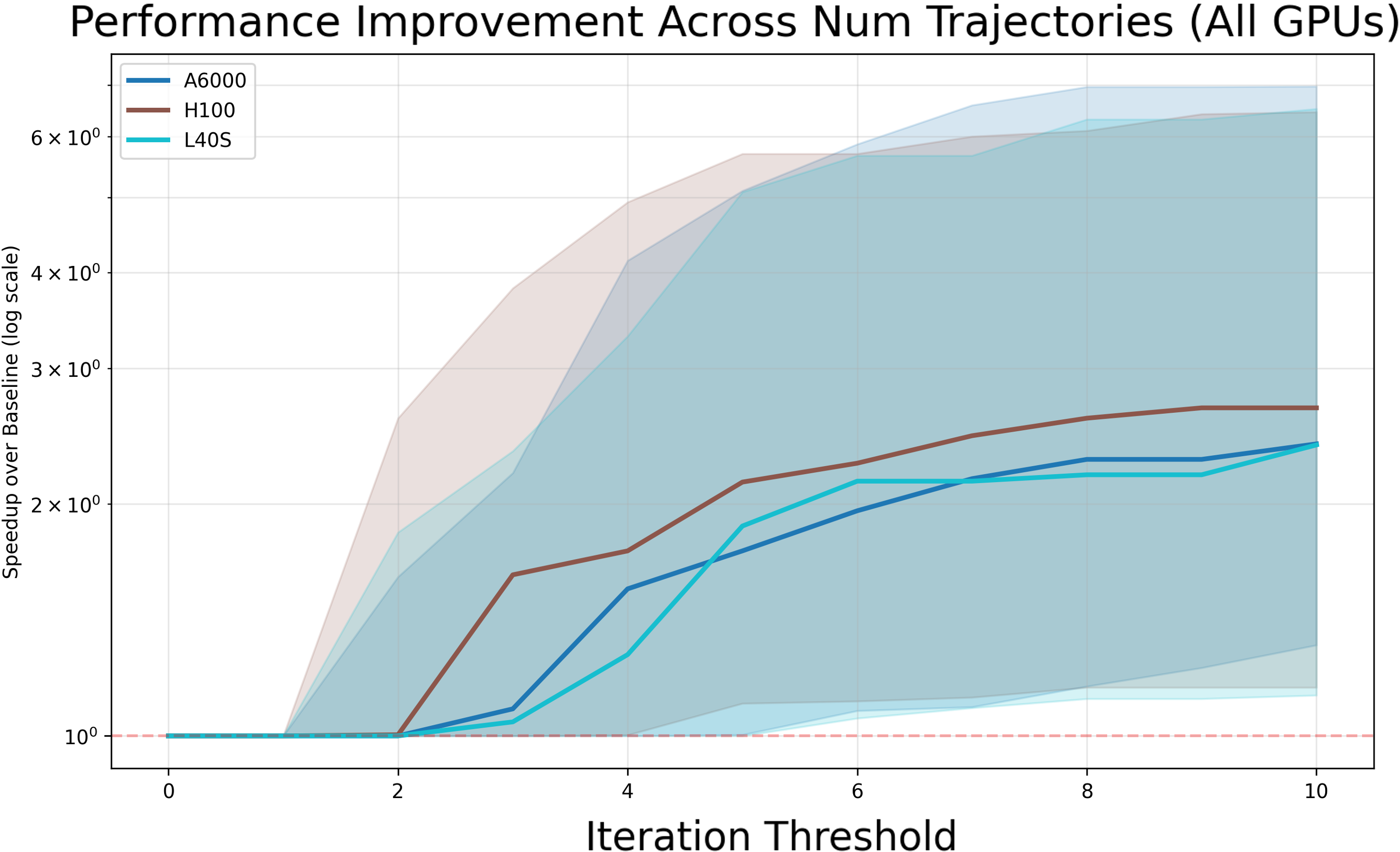}
        \caption{Performance vs.\ trajectory count (search breadth).}
        \label{fig:Num-Traj}
    \end{subfigure}
    \hfill
    \begin{subfigure}[t]{0.52\linewidth}
        \centering
        \includegraphics[width=\linewidth]{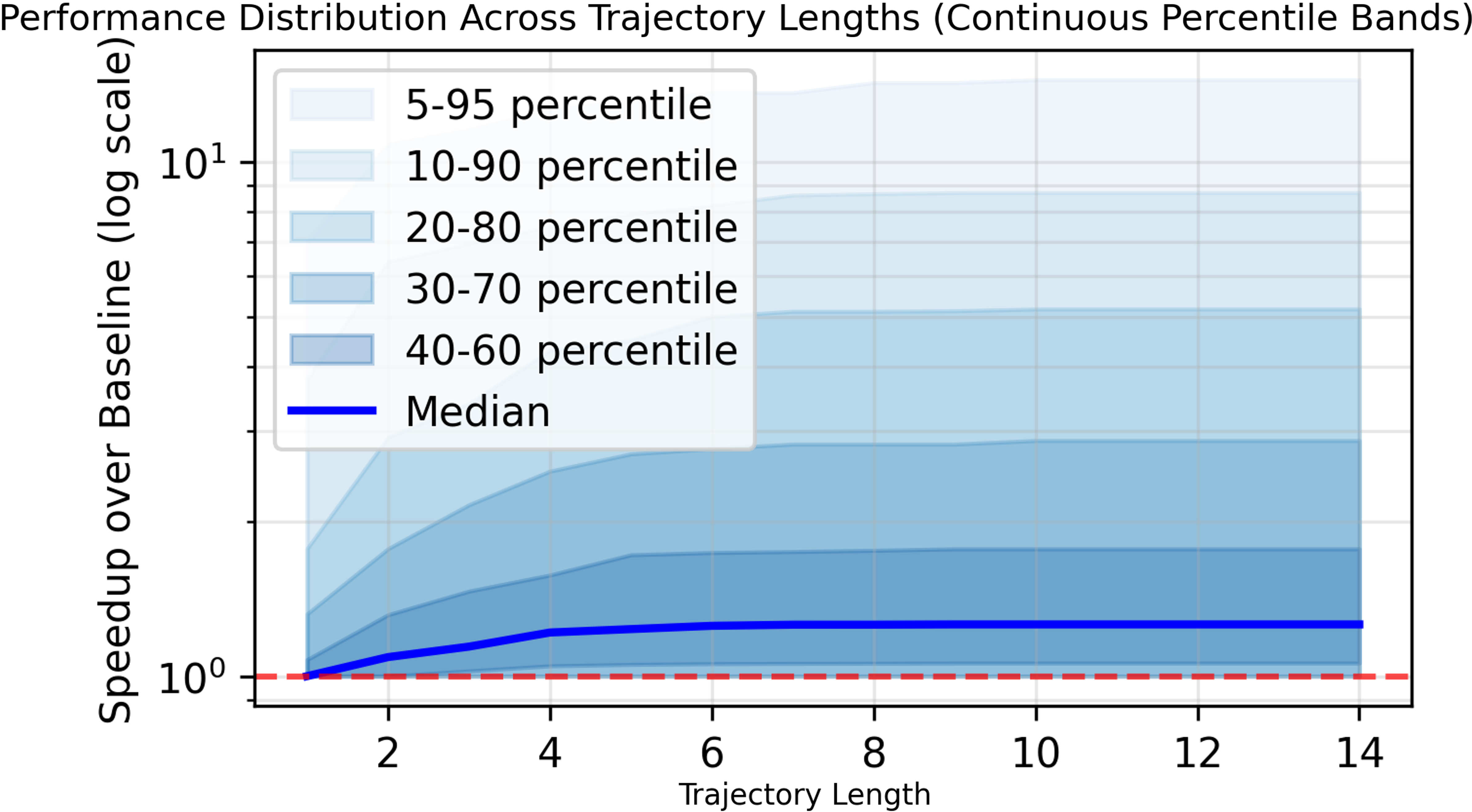}
        \caption{Performance vs.\ trajectory length (search depth).}
        \label{fig:Traj_Length}
    \end{subfigure}
    \caption{The impact of hyperparameter tuning on search dynamics: shaded regions show the inter-quartile range of optimized kernels. Breadth and depth both exhibit diminishing returns. }
    \label{fig:search_pair}
\end{figure}



\subsection{Cost and Efficiency}\label{app:budget}

Because token prices vary across model, date, and currency, we report token usage directly. The figures below come from the per-problem cost logs of 22 CUDA problems run with \texttt{gpt-5.1}. On KernelBench Levels~1 and~2, \sys consumes geometric means of 2,297,652 and 1,600,997 tokens per problem, respectively. Level~2 therefore costs 30\% fewer tokens, while the speedups of those runs were higher (2.50$\times$ against 1.43$\times$), because the Critic Reflection Agent ends trajectories earlier when improvements arrive quickly. Token count varies with code size, number of profiled kernels, and optimization complexity. Cost is tracked separately for kernels that were optimized successfully and kernels that were not, and failed attempts account for most of the wasted spend. In the current flow, the Level~3 headline sweep on the L40S (48 problems) cost about \$3,500 in LLM spend at list prices, about \$73 per problem. A further metric is speedup per persistent learning signal, since each rollout contributes scalar outcomes and textual critiques that later tasks can reuse.

Quality depends on the search budget more weakly than the default budget suggests. The median speedup saturates at about four transformations per trajectory, so the default depth of ten steps is roughly 2.5 times deeper than necessary, and a configuration of 8 trajectories with 4 steps each retains most of the median speedup of the full budget at about 32\% of the tokens (see the search breadth and depth analysis above). Relative to a minimal agent that receives the same CUDA code and NCU data, \sys uses 2.4 times fewer tokens and achieves about 2.6 times more performance improvement per token (Appendix~\ref{app:minimal_agent}).

\subsection{Optimization Distribution on Secondary HW Targets}

\begin{figure*}[h]
    \centering
        \centering
        \includegraphics[width=0.99\linewidth]{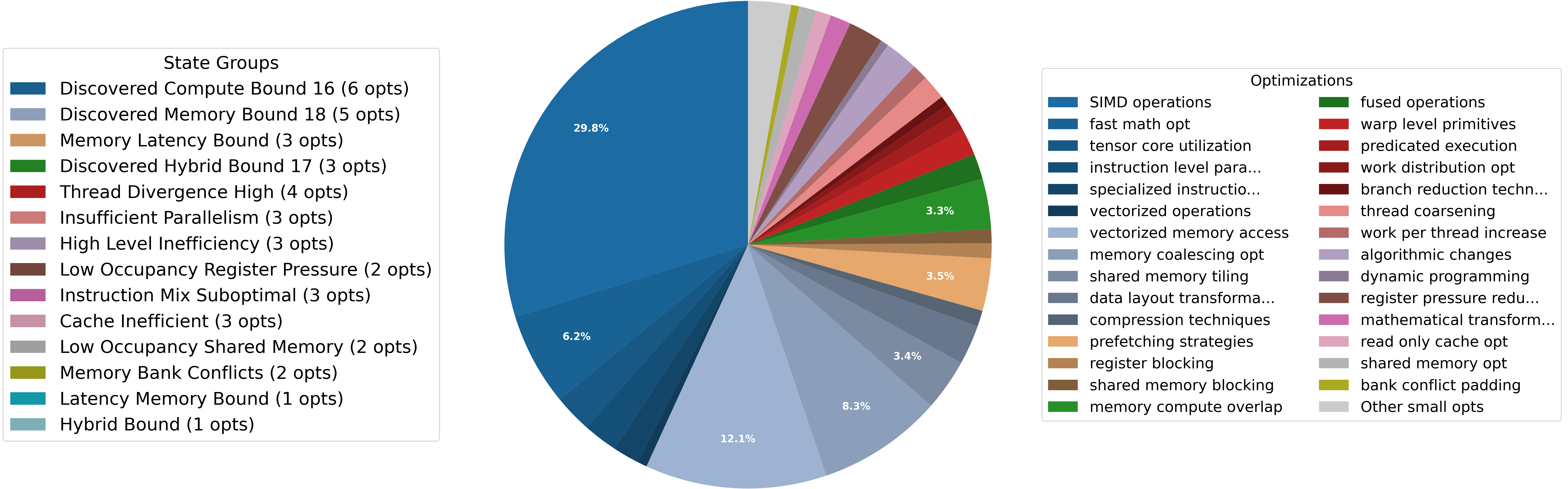}
        \caption{Distribution of Qualcomm optimization applications across performance states (KernelBench L1, Adreno 650 GPU).}
        \label{fig:opt_pie_qualcomm}
\end{figure*}

In addition to characterization the optimization distribution on the primary NVIDIA GPU targets, we also provide a summary of the optimization distribution on Qualcomm targets in Figure~\ref{fig:opt_pie_qualcomm}. Compared to NVIDIA GPUs, we observe a smaller set of optimizations dominating total optimization usage. This is due to the fact that the OpenCL compilation process requires greater hand tuning, whereas certain optimizations can be implemented directly by the CUDA compilation process. Additionally, we observe certain CUDA specific optimizations selected in the Qualcomm flow (such as tensor core utilization). While these are initially discovered and attempted, they do not result in code with improved performance, and no longer dominate optimization attempts.

\subsection{Optimization/State Clustering Analysis}

\begin{figure}
    \centering
    \includegraphics[width=\linewidth]{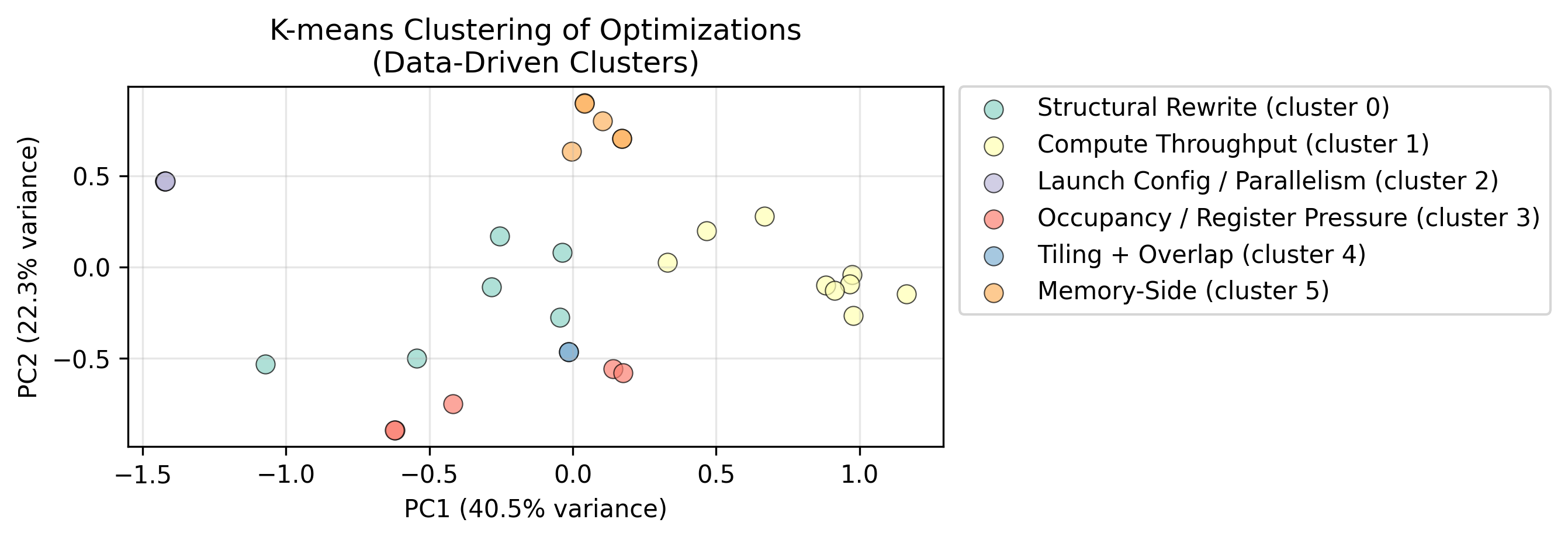}
    \caption{Synthetic Clustering of Top Optimizations}
    \label{fig:kmeans}
\end{figure}

We use the trajectories stored in \sys's \KB to validate the state abstraction used by MAIC-RL. Each successful CUDA action is represented by its pre-application profiler state, a 15-dimensional Nsight Compute bottleneck signature covering memory and compute throughput, occupancy, cache behavior, pipeline utilization, and warp scheduling. K-means clustering over these signatures recovers six interpretable action regimes: memory-side fixes, launch/parallelism fixes, occupancy and register-pressure interventions, tiling with compute-memory overlap, instruction-level compute tuning, and structural rewrites. This state-conditioned view is substantially more informative than clustering actions by usage frequency or binary success vectors, which mostly separate outliers from the rest. The clusters also align with the \KB's manual state taxonomy: memory-bandwidth states select memory-side actions, compute-throughput states select instruction-level tuning, balanced memory-compute states select tiling/overlap, and insufficient-parallelism states select launch-configuration changes. The only heterogeneous regime is structural rewrite, whose pre-action signatures are saturated across many metrics, consistent with its role as a last-resort intervention. This analysis suggests that \sys should retrieve and update actions through profiler-derived macro-states, while allowing the critic to refine or merge finer manual states when trajectory evidence does not support them.

\subsection{Correctness, Verification, and Cheating Prevention}\label{app:correctness}
To prevent shortcut behaviors such as calling PyTorch or emitting non-vanilla-CUDA code, we employ a soft-verification agent that compares the optimized kernel against the original KernelBench baseline. In addition to our test harness checking numerical correctness, we use an LLM agent that detects reward hacking (such as eliminating or moving operations out of profiled regions) and calling external optimized libraries, directly addressing concerns about cheating in agent-generated kernels. Finally, we verifiy verification agent's robustness by measuring its agreement with expert human reviewers. 

However, the protocol does not constitute a proof of behavioral equivalence and has several limitations that should be acknowledged. First, the verification agent is itself an LLM and inherits the underlying model's reliability characteristics: it can produce both false positives (rejecting legitimate optimizations whose surface form resembles a known cheating pattern, such as algebraic simplifications that genuinely eliminate redundant operations) and false negatives (missing novel evasion strategies that do not match the prompt's enumerated cheat templates). Second, the cheat-pattern catalog is anchored to the KernelBench/CUDA setting (e.g., dispatching to PyTorch, calling cuDNN/cuBLAS, hoisting operations out of the timed region); the OpenCL/Adreno and RVV/RISC-V backends use the same conceptual checks, but their cheat patterns have not been independently audited and may differ. Fourth, compiler-level shortcuts are outside the agent's view: if the emitted CUDA is structured such that NVCC, the driver, or a vendor library substitutes a fast-path implementation internally, the verification agent does not observe that substitution. 

Finally, a stronger protocol would combine LLM-level pattern detection with deterministic static analysis (e.g., checking the call graph for forbidden symbols) and randomized differential testing across broader input distributions. We document these limitations explicitly so that the verification step is auditable rather than represented as eliminating reward hacking entirely.

\subsection{Limitations}
\sys's system also exposes limitations for future work. Full-model optimization requires hierarchical decomposition because whole-network code generation introduces long boilerplate and heterogeneous performance states. Cross-target optimization requires stronger profiling abstractions so that bottlenecks can be compared across vendors and execution models. Persistent memories produced by many agents could also become high-signal datasets for training future models, converting short-term in-context adaptation into durable model-level capability. More broadly, \sys suggests a path toward optimization agents that learn from deployment feedback, retain domain knowledge across tasks, and adapt to new hardware targets without retraining.

\section{Representative Code Samples}\label{app:code_samples}
\subsection{\textbf{Algebraic Fusion:} (KernelBench Level~2, problem 72)}\label{app:algebraic_example}

In KernelBench Level~2 problem 72 (ConvTranspose3d, BatchNorm, AvgPool, AvgPool), the whole model in the L40S sweep takes 33.8\,$\mu$s against 10,632\,$\mu$s for PyTorch eager (315$\times$), 10,620\,$\mu$s for a CUDA graph, and 8,406\,$\mu$s for \texttt{torch.compile} (249$\times$). The seed implementation takes 33.3\,ms. The model applies a transposed 3D convolution (kernel 3, stride 2, padding 1), a BatchNorm, and two $2\times2\times2$ average pools. With BatchNorm folded to a per-channel scale and shift, every step after the convolution is linear, so the pooled output equals a $3\times3\times3$ stride-2 convolution of the input with folded weights. The kernel computes that fold directly. For each of the 81 pairs of input channel and tap and each of the 16 output channels, it sums the transposed-convolution taps that land in the same $4\times4\times4$ pooling window, applies the BatchNorm scale, divides by 64, and keeps the result in shared memory. Each thread then computes one output position for all 16 channels in fp32 and adds the folded bias and shift. The rewrite is exact up to fp16 rounding, and it holds when BatchNorm runs in inference mode, which is how both the flow and the eager baseline evaluate the model. The 34\,$\mu$s run reads about 13\,MB and writes about 7\,MB.

\begin{lstlisting}[style=cppstyle]
// KernelBench Level 2, problem 72 (ConvTranspose3d, BatchNorm, AvgPool, AvgPool), L40S sweep, final slot
#include <cuda_fp16.h>
#include <stdint.h>

// Fused slot: conv_transpose(+bias) -> BN(scale,shift) -> avgpool(2) -> avgpool(2).
// Per output (n, oc, z, y, x), z,y,x in [0,15):
//   out = scale[oc] * sum_{ic, a,b,c in 0..2} x[n,ic,2z+a,2y+b,2x+c] * Gs[ic,oc,a,b,c]
//         + (scale[oc]*bias[oc] + shift[oc])
// Gs = (1/64) * sum of convT weights over taps landing in the 4x4x4 pool window.
// Each thread computes one spatial output position for all 16 channels in fp32.

__global__ void __launch_bounds__(256)
k_fused(const __half* __restrict__ x, const __half* __restrict__ w,
        const __half* __restrict__ bias, const __half* __restrict__ scale,
        const __half* __restrict__ shift, __half* __restrict__ out) {
    __shared__ __align__(16) float Gs[81][16];   // [ic*27 + a*9 + b*3 + c][oc], scale folded in
    __shared__ float cst[16];                    // scale*bias + shift

    const int tid = threadIdx.x;

    // ---- fold weights for the 81 effective input taps ----
    for (int idx = tid; idx < 81 * 16; idx += 256) {
        const int kg = idx >> 4;
        const int oc = idx & 15;
        const int ic = kg / 27;
        const int t = kg - ic * 27;
        const int a = t / 9;
        const int b = (t / 3) % 3;
        const int c = t % 3;
        const int alo = (a == 0) ? 1 : 0, ahi = (a == 2) ? 0 : 2;
        const int blo = (b == 0) ? 1 : 0, bhi = (b == 2) ? 0 : 2;
        const int clo = (c == 0) ? 1 : 0, chi = (c == 2) ? 0 : 2;
        const __half* wp = w + (size_t)(ic * 16 + oc) * 27;
        float s = 0.f;
        for (int kz = alo; kz <= ahi; kz++)
            for (int ky = blo; ky <= bhi; ky++)
                for (int kx = clo; kx <= chi; kx++)
                    s += __half2float(__ldg(wp + kz * 9 + ky * 3 + kx));
        const float sc = __half2float(scale[oc]);
        Gs[kg][oc] = s * (1.0f / 64.0f) * sc;
    }
    if (tid < 16) {
        cst[tid] = __half2float(scale[tid]) * __half2float(bias[tid]) + __half2float(shift[tid]);
    }
    __syncthreads();

    const int gid = blockIdx.x * 256 + tid;
    const int total = 64 * 15 * 15 * 15;
    if (gid >= total) return;

    // decode (n, z, y, x)
    const int n = gid / 3375;
    int r = gid - n * 3375;
    const int z = r / 225;
    const int r2 = r - z * 225;
    const int y = r2 / 15;
    const int xx = r2 - y * 15;

    float acc[16];
#pragma unroll
    for (int oc = 0; oc < 16; oc++) acc[oc] = 0.f;

    const __half2* x2 = reinterpret_cast<const __half2*>(x);

#pragma unroll
    for (int ic = 0; ic < 3; ic++) {
#pragma unroll
        for (int a = 0; a < 3; a++) {
#pragma unroll
            for (int b = 0; b < 3; b++) {
                // row start (in elements) for input plane (n,ic), depth 2z+a, row 2y+b; col 2xx..2xx+2
                const int rowbase = ((n * 3 + ic) * 32768 + (2 * z + a) * 1024 + (2 * y + b) * 32) >> 1;
                const __half2 p0 = __ldg(x2 + rowbase + xx);       // cols 2xx, 2xx+1
                const __half2 p1 = __ldg(x2 + rowbase + xx + 1);   // cols 2xx+2, 2xx+3
                const float2 f0 = __half22float2(p0);
                const float v[3] = { f0.x, f0.y, __low2float(p1) };
#pragma unroll
                for (int c = 0; c < 3; c++) {
                    const int kg = ic * 27 + a * 9 + b * 3 + c;
                    const float vc = v[c];
#pragma unroll
                    for (int q = 0; q < 4; q++) {
                        const float4 g = *reinterpret_cast<const float4*>(&Gs[kg][q * 4]);
                        acc[q * 4 + 0] = fmaf(vc, g.x, acc[q * 4 + 0]);
                        acc[q * 4 + 1] = fmaf(vc, g.y, acc[q * 4 + 1]);
                        acc[q * 4 + 2] = fmaf(vc, g.z, acc[q * 4 + 2]);
                        acc[q * 4 + 3] = fmaf(vc, g.w, acc[q * 4 + 3]);
                    }
                }
            }
        }
    }

    const size_t obase = (size_t)n * 16 * 3375 + (size_t)z * 225 + y * 15 + xx;
#pragma unroll
    for (int oc = 0; oc < 16; oc++) {
        out[obase + (size_t)oc * 3375] = __float2half_rn(acc[oc] + cst[oc]);
    }
}

void kernel_region__all__9d141d8c2c_f16(const __half* in_x, const __half* w_conv_transpose_weight, const __half* w_conv_transpose_bias, __half* scratch_conv_transpose, const __half* in_batch_norm__scale, __half* scratch_batch_norm__g1, const __half* in_batch_norm__shift, __half* scratch_batch_norm, __half* scratch_avg_pool1, __half* out_avg_pool2) {
    const int total = 64 * 15 * 15 * 15;
    const int blocks = (total + 255) / 256;
    k_fused<<<blocks, 256>>>(in_x, w_conv_transpose_weight, w_conv_transpose_bias,
                             in_batch_norm__scale, in_batch_norm__shift, out_avg_pool2);
}
\end{lstlisting}

\subsection{\textbf{Tensor Core GEMM with Fused Epilogue:} (KernelBench Level~2, problem 37)}\label{app:gemm_example}

In KernelBench Level~2 problem 37 (Matmul, Swish, Sum, GroupNorm), the final model takes 1,897\,$\mu$s on the L40S, against 6,048\,$\mu$s for PyTorch eager (3.19$\times$), 5,998\,$\mu$s for a CUDA graph, and 2,170\,$\mu$s for \texttt{torch.compile} (1.14$\times$). The problem multiplies a $32768\times1024$ input by a $1024\times4096$ weight matrix and applies a bias, Swish, a second bias, and GroupNorm. The kernel computes the GEMM with $128\times128$ block tiles and eight warps, two along $M$ and four along $N$, each owning a $64\times32$ tile. Operand tiles stream through a three-stage \texttt{cp.async} pipeline into XOR-swizzled shared memory, are loaded with \texttt{ldmatrix}, and are multiplied with \texttt{mma.sync} m16n8k16 instructions on fp16 inputs with fp32 accumulation. The epilogue never writes the intermediate matrix to global memory. It adds the bias, applies the sigmoid and the multiplication of Swish with intermediate values rounded to fp16, adds the second bias, stages the tile in shared memory, computes GroupNorm over each 64-channel group with one thread per row and group, and writes the result with coalesced 16-byte stores. The listing is highlighted: small helper wrappers, the accumulator initialization, and the launch-attribute setup are omitted.

\begin{lstlisting}[style=cppstyle]
// KernelBench Level 2, problem 37 (Matmul, Swish, Sum, GroupNorm), L40S sweep, final slot; highlighted
// (cp.async, commit/wait and rounding helpers, accumulator initialization and launch-attribute setup are omitted)
#define BM 128
#define BN 128
#define BK 32
#define LDS 32
#define STG 3
#define KD 1024
#define ND 4096
#define MD 32768
#define LDCS 136
#define THR 256

__device__ __forceinline__ void ldm4(uint32_t* r, uint32_t a) {
    asm volatile("ldmatrix.sync.aligned.m8n8.x4.shared.b16 {%0,%1,%2,%3}, [%4];\n"
                 : "=r"(r[0]), "=r"(r[1]), "=r"(r[2]), "=r"(r[3]) : "r"(a));
}
__device__ __forceinline__ void mma(float* c, const uint32_t* a, uint32_t b0, uint32_t b1) {
    asm volatile("mma.sync.aligned.m16n8k16.row.col.f32.f16.f16.f32 {%0,%1,%2,%3}, {%4,%5,%6,%7}, {%8,%9}, {%0,%1,%2,%3};\n"
                 : "+f"(c[0]), "+f"(c[1]), "+f"(c[2]), "+f"(c[3])
                 : "r"(a[0]), "r"(a[1]), "r"(a[2]), "r"(a[3]), "r"(b0), "r"(b1));
}

// 128x128 block tile, 8 warps (2 along M x 4 along N), each warp 64x32.
__global__ void __launch_bounds__(THR, 2)
k_fused(const __half* __restrict__ X, const __half* __restrict__ W, const __half* __restrict__ Bl,
        const __half* __restrict__ Bi, const __half* __restrict__ gam, const __half* __restrict__ bet,
        __half* __restrict__ Y) {
    extern __shared__ __align__(16) unsigned char smem[];
    __half* sA = (__half*)smem;                       // STG * BM * LDS
    __half* sB = sA + STG * BM * LDS;                 // STG * BN * LDS
    const int tid = threadIdx.x, lane = tid & 31, wid = tid >> 5;
    const int wm = wid & 1, wn = wid >> 1;            // wn in 0..3
    const int nt = blockIdx.x & 31, mt = blockIdx.x >> 5;
    const __half* Xb = X + (size_t)mt * BM * KD;
    const __half* Wb = W + (size_t)nt * BN * KD;

    auto load = [&](int st, int kt) {
        __half* a = sA + st * BM * LDS;
        __half* b = sB + st * BN * LDS;
        const int k0 = kt * BK;
#pragma unroll
        for (int i = 0; i < 2; i++) {
            const int c = tid + i * THR;
            const int r = c >> 2, cc = (c & 3) * 8;
            cp16(a + r * LDS + ((((c & 3) ^ ((r >> 1) & 3))) << 3), Xb + (size_t)r * KD + k0 + cc);
        }
#pragma unroll
        for (int i = 0; i < 2; i++) {
            const int c = tid + i * THR;
            const int r = c >> 2, cc = (c & 3) * 8;
            cp16(b + r * LDS + ((((c & 3) ^ ((r >> 1) & 3))) << 3), Wb + (size_t)r * KD + k0 + cc);
        }
    };


    // ... accumulator tile acc[4][4][4] zero-initialized

    const int KT = KD / BK;
    load(0, 0); cpc();
    load(1, 1); cpc();
    for (int kt = 0; kt < KT; kt++) {
        cpw<1>();
        __syncthreads();
        if (kt + 2 < KT) load((kt + 2) % STG, kt + 2);
        cpc();
        const __half* a = sA + (kt % STG) * BM * LDS;
        const __half* b = sB + (kt % STG) * BN * LDS;
#pragma unroll
        for (int kk = 0; kk < 2; kk++) {
            uint32_t af[4][4], bf[2][4];
#pragma unroll
            for (int i = 0; i < 4; i++) {
                const int r = wm * 64 + i * 16 + (lane & 15);
                const int ch = (kk * 2 + (lane >> 4)) ^ ((r >> 1) & 3);
                ldm4(af[i], su32(a + r * LDS + ch * 8));
            }
#pragma unroll
            for (int j = 0; j < 2; j++) {
                const int r = wn * 32 + j * 16 + (lane & 7) + ((lane >> 4) << 3);
                const int ch = (kk * 2 + ((lane >> 3) & 1)) ^ ((r >> 1) & 3);
                ldm4(bf[j], su32(b + r * LDS + ch * 8));
            }
#pragma unroll
            for (int i = 0; i < 4; i++)
#pragma unroll
                for (int j = 0; j < 2; j++) {
                    mma(acc[i][2 * j], af[i], bf[j][0], bf[j][1]);
                    mma(acc[i][2 * j + 1], af[i], bf[j][2], bf[j][3]);
                }
        }
    }
    cpw<0>();
    __syncthreads();

    // epilogue 1: bias, sigmoid, mul, add -> smem (fp16)
    __half* sC = (__half*)smem;
#pragma unroll
    for (int j = 0; j < 4; j++) {
        const int col = wn * 32 + j * 8 + (lane & 3) * 2;
        const int gc = nt * BN + col;
        float2 bl = __half22float2(*(const __half2*)(Bl + gc));
        float2 bi = __half22float2(*(const __half2*)(Bi + gc));
#pragma unroll
        for (int i = 0; i < 4; i++) {
#pragma unroll
            for (int h = 0; h < 2; h++) {
                const int row = wm * 64 + i * 16 + (lane >> 2) + h * 8;
                float v0 = rh(acc[i][j][2 * h] + bl.x), v1 = rh(acc[i][j][2 * h + 1] + bl.y);
                float s0 = rh(__fdividef(1.0f, 1.0f + __expf(-v0)));
                float s1 = rh(__fdividef(1.0f, 1.0f + __expf(-v1)));
                float m0 = rh(s0 * v0), m1 = rh(s1 * v1);
                *(__half2*)(sC + row * LDCS + col) = __floats2half2_rn(m0 + bi.x, m1 + bi.y);
            }
        }
    }
    __syncthreads();

    // epilogue 2: group norm per (row, 64-channel group), in place in smem.
    // 256 threads cover 128 rows x 2 groups in one pass.
    {
        const int row = tid >> 1, g = tid & 1;
        float v[64];
#pragma unroll
        for (int c = 0; c < 8; c++) {
            const int cr = (c + 2 * g) & 7;
            uint4 q = *(const uint4*)(sC + row * LDCS + g * 64 + cr * 8);
            const __half2* hq = (const __half2*)&q;
#pragma unroll
            for (int e = 0; e < 4; e++) {
                float2 f = __half22float2(hq[e]);
                v[c * 8 + 2 * e] = f.x; v[c * 8 + 2 * e + 1] = f.y;
            }
        }
        float s = 0.f;
#pragma unroll
        for (int e = 0; e < 64; e++) s += v[e];
        const float mean = s * (1.0f / 64.0f);
        float qv = 0.f;
#pragma unroll
        for (int e = 0; e < 64; e++) { float d = v[e] - mean; qv += d * d; }
        const float inv = rsqrtf(qv * (1.0f / 64.0f) + 1e-5f);
#pragma unroll
        for (int c = 0; c < 8; c++) {
            const int cr = (c + 2 * g) & 7;
            const int ch = nt * BN + g * 64 + cr * 8;
            uint4 gq = __ldg((const uint4*)(gam + ch));
            uint4 bq = __ldg((const uint4*)(bet + ch));
            const __half2* gh = (const __half2*)&gq;
            const __half2* bh = (const __half2*)&bq;
            uint4 o; __half2* oh = (__half2*)&o;
#pragma unroll
            for (int e = 0; e < 4; e++) {
                float2 gg = __half22float2(gh[e]), bb = __half22float2(bh[e]);
                float y0 = (v[c * 8 + 2 * e] - mean) * inv * gg.x + bb.x;
                float y1 = (v[c * 8 + 2 * e + 1] - mean) * inv * gg.y + bb.y;
                oh[e] = __floats2half2_rn(y0, y1);
            }
            *(uint4*)(sC + row * LDCS + g * 64 + cr * 8) = o;
        }
    }
    __syncthreads();

    // coalesced store
#pragma unroll 1
    for (int i = tid; i < BM * 16; i += THR) {
        const int r = i >> 4, cc = i & 15;
        uint4 q = *(const uint4*)(sC + r * LDCS + cc * 8);
        *(uint4*)(Y + (size_t)(mt * BM + r) * ND + nt * BN + cc * 8) = q;
    }
}

void kernel_region__all__6723b6a1a9_f16(const __half* in_x, const __half* w_matmul_weight, const __half* w_matmul_bias, __half* scratch_matmul, __half* scratch_sigmoid, __half* scratch_mul, const __half* in_bias, __half* scratch_add, const __half* w_group_norm_weight, const __half* w_group_norm_bias, __half* out_group_norm) {
    // ... dynamic shared-memory size and launch attribute setup omitted
    k_fused<<<(MD / BM) * (ND / BN), THR, smem>>>(in_x, w_matmul_weight, w_matmul_bias, in_bias,
                                                  w_group_norm_weight, w_group_norm_bias, out_group_norm);
}
\end{lstlisting}

\subsection{Level~3 Full Models}

\label{sec:appendix_level3}

To better understand how \sys behaves on end-to-end workloads, we analyze several representative KernelBench Level~3 problems in detail. Table~\ref{tab:l3_models} gives the speedups of these models in the L40S sweep behind Table~\ref{tab:summary_all}, over PyTorch eager.

\begin{table}[h]
\centering
\small
\caption{Speedup of \sys on representative KernelBench Level~3 models on an L40S, over PyTorch eager. The last row is the geometric mean over all 46 problems.}
\label{tab:l3_models}
\begin{tabular}{@{}lr@{}}
\toprule
Model & Speedup \\
\midrule
LeNet5 & 10.58$\times$ \\
SwinMLP & 4.07$\times$ \\
ResNetBasicBlock & 3.19$\times$ \\
DenseNet121DenseBlock & 2.72$\times$ \\
SqueezeNetFireModule & 2.41$\times$ \\
MLP & 1.07$\times$ \\
\midrule
All 46 problems (geometric mean) & 5.18$\times$ \\
\bottomrule
\end{tabular}
\end{table}

\paragraph{LeNet5.}
For the Level~3 network \textit{LeNet5}, the final model takes 51.2\,$\mu$s on the L40S against 5,712\,$\mu$s for its seed, and runs 10.6$\times$ faster than PyTorch eager. The kernel below fuses the second convolution, the ReLU, and the $2\times2$ max pool into one tensor-core kernel. The convolution (6 input channels, 16 output channels, $5\times5$ filters) is lowered to a GEMM in which each row is an image position at an even output column. The six-wide input window of that row feeds the even output column and, with the weights shifted by one, the next odd column, so the GEMM has $N=32$ (16 channels times two column parities) and $K=180$ padded to 192. The horizontal part of the pooling is a maximum of two accumulator fragments, and the vertical part is a warp shuffle between lanes that hold adjacent rows. The weights stay in registers as MMA fragments, the inputs arrive through \texttt{cp.async} into double-buffered chunks of eight images, and persistent blocks loop over the chunks. The listing is highlighted: the helper that counts SMs, the naive fallback kernels, and the shape check are omitted.

\begin{lstlisting}[style=cppstyle]
// KernelBench Level 3, 4_LeNet5, L40S sweep, final fused conv + ReLU + maxpool slot (variant v1); highlighted
// (the SM-count helper, the naive fallback kernels, and the shape-dependent dispatch details are omitted)
// conv(16x6x5x5,s1) + relu + maxpool(2x2,s2), fp16 io / fp32 acc on tensor cores.
// GEMM row = (img, oh, ow_e even); its 6-wide input window (3 aligned words per
// (ic,kh)) feeds both ow_e (even weight set, kw 0..4) and ow_e+1 (odd weight set,
// shifted by one) -> N = 32, K = 6*5*6 = 180 (pad 192). A loads are 32-bit, aligned.
// Persistent blocks, cp.async double-buffered 8-image chunks.

__device__ __forceinline__ void fz_mma(float *c, const unsigned *a, unsigned b0, unsigned b1) {
    asm volatile(
        "mma.sync.aligned.m16n8k16.row.col.f32.f16.f16.f32 {%0,%1,%2,%3}, {%4,%5,%6,%7}, {%8,%9}, {%0,%1,%2,%3};\n"
        : "+f"(c[0]), "+f"(c[1]), "+f"(c[2]), "+f"(c[3])
        : "r"(a[0]), "r"(a[1]), "r"(a[2]), "r"(a[3]), "r"(b0), "r"(b1));
}
__device__ __forceinline__ void fz_cp16(unsigned sa, const void *src) {
    asm volatile("cp.async.cg.shared.global [%0], [%1], 16;\n" :: "r"(sa), "l"(src));
}

#define FZ_IMGS 8
#define FZ_WORDS (FZ_IMGS * 588)

__device__ __forceinline__ int fz_rowbase(int R) {
    int img = R / 50, lr = R - img * 50;
    int ph = lr / 10, rem = lr - ph * 10;
    int pw = rem >> 1, ohb = rem & 1;
    int oh = 2 * ph + ohb;
    return img * 588 + oh * 7 + pw;
}

__global__ void __launch_bounds__(128, 2)
k_fused_fast(const __half *__restrict__ x, const __half *__restrict__ w, const __half *__restrict__ bias,
             __half *__restrict__ y, int nch) {
    __shared__ __align__(16) unsigned sIn[2][FZ_WORDS];
    __shared__ __align__(16) __half sOut[FZ_IMGS * 400];
    __shared__ int2 sOff2[48];

    const int tid = threadIdx.x, lane = tid & 31, warp = tid >> 5;
    const int g = lane >> 2, t = lane & 3;

    int chunk = blockIdx.x;
    if (chunk >= nch) return;

    const unsigned sBase = (unsigned)__cvta_generic_to_shared(&sIn[0][0]);
    auto load = [&](int ch, int st) {
        const char *src = reinterpret_cast<const char *>(x + (long)ch * (FZ_IMGS * 1176));
        unsigned dst = sBase + st * (FZ_WORDS * 4);
        for (int i = tid; i < FZ_WORDS / 4; i += 128) fz_cp16(dst + i * 16, src + (long)i * 16);
    };
    load(chunk, 0);
    asm volatile("cp.async.commit_group;\n" ::);

    // word offset table: kwd -> ic*98 + kh*7 + q
    if (tid < 48) {
        int ks = tid >> 2, tt = tid & 3;
        int o[2];
#pragma unroll
        for (int j = 0; j < 2; j++) {
            int kwd = ks * 8 + tt + 4 * j;
            int v = 0;
            if (kwd < 90) {
                int ic = kwd / 15, r = kwd - ic * 15, kh = r / 3, q = r - kh * 3;
                v = ic * 98 + kh * 7 + q;
            }
            o[j] = v;
        }
        sOff2[tid] = make_int2(o[0], o[1]);
    }

    // weight fragments
    unsigned bw[12][4][2];
#pragma unroll
    for (int ks = 0; ks < 12; ks++)
#pragma unroll
        for (int nt = 0; nt < 4; nt++) {
            int n = nt * 8 + g;
            int oc = n & 15, par = n >> 4;
#pragma unroll
            for (int j = 0; j < 2; j++) {
                unsigned pk = 0u;
#pragma unroll
                for (int h = 0; h < 2; h++) {
                    int k = ks * 16 + 2 * t + 8 * j + h;
                    int kwd = k >> 1;
                    unsigned hv = 0u;
                    if (kwd < 90) {
                        int ic = kwd / 15, r = kwd - ic * 15, kh = r / 3, q = r - kh * 3;
                        int p = 2 * q + h;
                        int kw = par ? p - 1 : p;
                        if (kw >= 0 && kw < 5)
                            hv = (unsigned)__half_as_ushort(w[((oc * 6 + ic) * 5 + kh) * 5 + kw]);
                    }
                    pk |= hv << (16 * h);
                }
                bw[ks][nt][j] = pk;
            }
        }
    float bv[2][2];
#pragma unroll
    for (int nt = 0; nt < 2; nt++) {
        int oc0 = nt * 8 + 2 * t;
        bv[nt][0] = bias ? __half2float(bias[oc0]) : 0.f;
        bv[nt][1] = bias ? __half2float(bias[oc0 + 1]) : 0.f;
    }

    for (int it = 0; chunk < nch; it++, chunk += gridDim.x) {
        const int s = it & 1;
        asm volatile("cp.async.wait_group 0;\n" ::);
        __syncthreads();
        int nxt = chunk + gridDim.x;
        if (nxt < nch) load(nxt, s ^ 1);
        asm volatile("cp.async.commit_group;\n" ::);

        const unsigned *S = sIn[s];
        for (int tile = warp; tile < 25; tile += 4) {
            const int R0 = tile * 16 + g;
            const int b0 = fz_rowbase(R0), b1 = fz_rowbase(R0 + 8);
            float c[4][4];
#pragma unroll
            for (int nt = 0; nt < 4; nt++) {
                c[nt][0] = bv[nt & 1][0]; c[nt][1] = bv[nt & 1][1];
                c[nt][2] = bv[nt & 1][0]; c[nt][3] = bv[nt & 1][1];
            }
#pragma unroll
            for (int ks = 0; ks < 12; ks++) {
                int2 o = sOff2[ks * 4 + t];
                unsigned a[4];
                a[0] = S[b0 + o.x];
                a[1] = S[b1 + o.x];
                a[2] = S[b0 + o.y];
                a[3] = S[b1 + o.y];
#pragma unroll
                for (int nt = 0; nt < 4; nt++) fz_mma(c[nt], a, bw[ks][nt][0], bw[ks][nt][1]);
            }
#pragma unroll
            for (int nt = 0; nt < 2; nt++)
#pragma unroll
                for (int e = 0; e < 4; e++) {
                    float v = fmaxf(c[nt][e], c[nt + 2][e]);
                    v = fmaxf(v, __shfl_xor_sync(0xffffffffu, v, 4));
                    v = fmaxf(v, 0.f);
                    if ((g & 1) == 0) {
                        int R = R0 + 8 * (e >> 1);
                        int img = R / 50, lr = R - img * 50;
                        int ph = lr / 10, rem = lr - ph * 10, pw = rem >> 1;
                        int oc = nt * 8 + 2 * t + (e & 1);
                        sOut[img * 400 + oc * 25 + ph * 5 + pw] = __float2half_rn(v);
                    }
                }
        }
        __syncthreads();
        uint4 *gy = reinterpret_cast<uint4 *>(y + (long)chunk * (FZ_IMGS * 400));
        const uint4 *sy = reinterpret_cast<const uint4 *>(sOut);
        for (int i = tid; i < FZ_IMGS * 400 / 8; i += 128) gy[i] = sy[i];
    }
}

// ... naive fallback kernels (k_gen_conv, k_gen_relu, k_gen_pool) omitted

void kernel_fused__conv2d_f16__relu_f16__maxpool2d_f16(const __half* a_input, const __half* a_weight, const __half* a_bias, __half* a_output, int a_N, int a_IC, int a_IH, int a_IW, int a_OC, int a_KH, int a_KW, int a_SH, int a_SW, int a_PH, int a_PW, int a_DH, int a_DW, __half* b_output, int b_n, __half* c_output, int c_N, int c_C, int c_IH, int c_IW, int c_KH, int c_KW, int c_SH, int c_SW, int c_PH, int c_PW, int c_DH, int c_DW) {
    // ... shape check against the LeNet5 conv2 + pool configuration (fast) omitted
    if (fast) {
        int nch = a_N / FZ_IMGS;
        int grid = fz_sm_count() * 2;
        if (grid > nch) grid = nch;
        k_fused_fast<<<grid, 128>>>(a_input, a_weight, a_bias, c_output, nch);
        return;
    }
    // ... fallback launches omitted
}
\end{lstlisting}

\paragraph{DenseNet121DenseBlock.}
In the DenseNet121 dense block, the final model takes 4,016\,$\mu$s against 315,792\,$\mu$s for its seed, and runs 2.72$\times$ faster than PyTorch eager. Dense blocks concatenate the outputs of earlier layers before every BatchNorm and ReLU. The kernel below fuses a two-input concatenation, the folded BatchNorm affine, and the ReLU into one pass. The concatenation costs no extra memory traffic: for each 16-byte chunk the index arithmetic selects the first or the second source tensor, so the concatenated tensor is never materialized. Each thread handles two chunks of eight fp16 values with block-strided indexing, so every warp access is coalesced and two loads are in flight before any compute. The per-channel scale and bias use the read-only path, and the streamed activations use evict-first loads and stores. The listing omits the scalar fallback kernel and the host-side size and alignment checks.

\begin{lstlisting}[style=cppstyle]
// KernelBench Level 3, 14_DenseNet121DenseBlock, L40S sweep, final fused concat + BatchNorm + ReLU slot; highlighted
// (the scalar fallback kernel and the alignment and size checks of the host code are omitted)
// Fused cat2 -> batchnorm2d (folded affine) -> relu, single pass.
// Vectorised: each thread handles two 16-byte (8 x fp16) chunks, block-strided
// (chunk base+t and base+256+t) so every warp instruction is fully coalesced and
// two 16B loads are in flight per thread before any compute/store.
// Per-channel scale/bias go through the read-only path (__ldg, tiny & reused);
// bulk streamed data uses evict-first loads/stores.
// Intermediates (a_output, b_output) are scratch and not materialised.

__device__ __forceinline__ uint4 mb_fuse8(uint4 v, float sc, float bb) {
    uint4 o;
    const __half2 *h = reinterpret_cast<const __half2 *>(&v);
    __half2 *oh = reinterpret_cast<__half2 *>(&o);
#pragma unroll
    for (int k = 0; k < 4; k++) {
        float2 f = __half22float2(h[k]);
        f.x = fmaxf(fmaf(sc, f.x, bb), 0.0f);
        f.y = fmaxf(fmaf(sc, f.y, bb), 0.0f);
        oh[k] = __floats2half2_rn(f.x, f.y);
    }
    return o;
}

__device__ __forceinline__ const uint4 *mb_src(const uint4 *__restrict__ a, const uint4 *__restrict__ b,
                                               unsigned v, unsigned Ca, unsigned Cb, unsigned C,
                                               unsigned HW, unsigned &c) {
    unsigned e = v * 8u;
    unsigned rowsz = C * HW;
    unsigned n = e / rowsz;
    unsigned rem = e - n * rowsz;
    c = rem / HW;
    unsigned p = rem - c * HW;
    if (c < Ca) {
        unsigned idx = n * Ca * HW + c * HW + p;
        return a + (idx >> 3);
    } else {
        unsigned idx = n * Cb * HW + (c - Ca) * HW + p;
        return b + (idx >> 3);
    }
}

__global__ void __launch_bounds__(256) k_fused_cat_bn_relu_vec2(
        const uint4 *__restrict__ a, const uint4 *__restrict__ b,
        uint4 *__restrict__ y, const __half *__restrict__ s, const __half *__restrict__ bi,
        unsigned nv, unsigned Ca, unsigned Cb, unsigned HW) {
    unsigned C = Ca + Cb;
    unsigned v0 = blockIdx.x * 512u + threadIdx.x;
    unsigned v1 = v0 + 256u;
    if (v0 >= nv) return;
    bool has1 = v1 < nv;
    unsigned c0, c1 = 0;
    const uint4 *p0 = mb_src(a, b, v0, Ca, Cb, C, HW, c0);
    const uint4 *p1 = has1 ? mb_src(a, b, v1, Ca, Cb, C, HW, c1) : p0;
    uint4 x0 = __ldcs(p0);
    uint4 x1 = make_uint4(0, 0, 0, 0);
    if (has1) x1 = __ldcs(p1);
    float s0 = __half2float(__ldg(s + c0)), b0 = __half2float(__ldg(bi + c0));
    __stcs(y + v0, mb_fuse8(x0, s0, b0));
    if (has1) {
        float s1 = __half2float(__ldg(s + c1)), b1 = __half2float(__ldg(bi + c1));
        __stcs(y + v1, mb_fuse8(x1, s1, b1));
    }
}

// ... scalar fallback kernel (k_fused_cat_bn_relu_scalar) omitted

void kernel_fused__cat2_c1_f16__batchnorm2d_f16__relu_f16(const __half* a_a, const __half* a_b, __half* a_output, int a_N, int a_H, int a_W, int a_Ca, int a_Cb, const __half* b_scale, const __half* b_bias, __half* b_output, int b_N, int b_C, int b_H, int b_W, __half* c_output, int c_n) {
    // ... size and alignment checks for the vectorized path omitted (vec)
    if (vec) {
        unsigned nv = (unsigned)(total / 8);
        unsigned blocks = (nv + 511) / 512;
        k_fused_cat_bn_relu_vec2<<<blocks, 256>>>(
            reinterpret_cast<const uint4 *>(a_a), reinterpret_cast<const uint4 *>(a_b),
            reinterpret_cast<uint4 *>(c_output), b_scale, b_bias,
            nv, (unsigned)Ca, (unsigned)Cb, (unsigned)HW);
    }
    // ... scalar fallback launch omitted
}
\end{lstlisting}

\paragraph{Observed Scaling Limitations.}
Despite these improvements, several limitations emerge when optimizing full-model workloads.

First, Level~3 models contain many kernels with diverse performance states and optimization opportunities. Because the current optimization pipeline processes a single optimization target per iteration, improvements to the overall model are limited by the sequential exploration process.

Second, implementing full networks directly in CUDA results in substantially larger source files compared to the concise Python representations used in KernelBench. The resulting verbosity introduces significant boilerplate code, which increases context length and reduces the clarity of performance-relevant signals for the LLM agent.

These observations suggest that future extensions of \sys should incorporate hierarchical preprocessing of the model graph, allowing the agent to operate on fused sub-blocks or groups of kernels. Given the performance improvements observed on Level~2 fused operators, such decomposition would likely improve optimization efficiency for full-model workloads.\\

\subsection{Qualcomm OpenCL Example}

We illustrate \sys's OpenCL optimization flow on KernelBench L1 problem 33, BatchNorm over a 4D activation tensor with 64 feature channels. Across ten parallel rollouts under the \texttt{gpt-5-mini-2025-08-07} policy, the best verified kernel runs in 294.52 ms on the Adreno 650 versus 1030.26 ms for the seed kernel, a 3.50x end-to-end speedup measured with warm-cache, post-warmup OpenCL event profiling. We use this example because it exercises the learned reduction-and-broadcast optimization template and represents a parallelism-strategy rewrite rather than a standard matmul tiling case.

The seed kernel maps one work-item to each output channel and performs three sequential scalar passes over the $N \times H \times W$ slice: mean accumulation, variance accumulation, and normalized affine writeback. This underutilizes the Adreno 650: only 64 work-items are launched, SIMD lanes and compute units remain mostly idle, global memory is accessed through scalar half-precision loads, and the channel slice is read three times.

The optimized kernel changes the mapping to one workgroup per channel. Work-items cooperatively accumulate both sum and sum-of-squares in one pass, compute variance using $\mathrm{Var}[x]=E[x^2]-E[x]^2$, reduce through a \texttt{\_\_local float2} buffer, broadcast the resulting mean and inverse standard deviation, and fuse normalization into the same kernel. The reduction loop uses \texttt{half8} vector loads for 128-bit global I/O, converts to \texttt{float8} for mixed-precision accumulation, and uses masked tail handling for non-multiple-of-eight slices. Shape constants such as \texttt{-DFEATURES=64} allow the OpenCL JIT to unroll and constant-fold indexing.

This kernel combines five recurring OpenCL optimization patterns discovered by \sys: vectorized 128-bit global I/O, cooperative \texttt{\_\_local} reduction, mixed-precision arithmetic, compile-time shape specialization, and explicit tail masking. Problem 33 is a compact example where all five appear together, and the emitted natural-language rationale in the optimized \texttt{.cl} file makes the hardware reasoning auditable.

\begin{lstlisting}[style=cppstyle]
#pragma OPENCL EXTENSION cl_khr_fp16 : enable

/*
 * batchnorm - widened vectorized implementation (half8 / float8)
 * - Uses half8 vload8/vstore8 for 128-bit global transactions
 * - Converts to float8 for math, splits into float4 pairs for dot/mad
 * - Tail handling builds a masked half8, does vector math, then scalar-stores only valid lanes
 * - Cooperative reduction in __local remains, barriers minimized to required locations
 *
 * Notes:
 * - Host should choose a local size that is a multiple of 32 for Adreno GPUs (wave size).
 * - Keep local_size <= 512 as guarded below.
 */

__kernel void batchnorm(
    __global const half* x,
    __global const half* weight,
    __global const half* bias,
    __global       half* y,
    const int            N,
    const int            H,
    const int            W)
{
    const int C = (int)get_num_groups(0); /* number of groups expected = channels */
    const int local_id = get_local_id(0);
    const int local_size = get_local_size(0);
    const int group_id = get_group_id(0); /* channel index (one group per channel) */
    const int HW = H * W;
    const int NHW = N * HW;

    /* Safety guards (as in original) */
    if (local_size > 512) return;
    if (group_id >= C) return;
    const int c = group_id;

    /* Vectorisation params: half8 / float8 (8 lanes = 128 bits) */
    const int VECTOR_LEN = 8; /* half8 / float8 lanes */
    const int TILE_STRIDE = local_size * VECTOR_LEN;

    /* local buffer: one float2 per work-item: .x = sum, .y = sumsq */
    __local float2 lacc[512];

    /* local params to broadcast mean, inv_mul (inv_std*weight), bias */
    __local float lparams[3]; /* [0]=mean, [1]=inv_mul, [2]=bias */

    const float4 ones4 = (float4)(1.0f, 1.0f, 1.0f, 1.0f);
    const float eps = 1e-5f;

    /* --- single pass: accumulate sum and sumsq per-thread --- */
    float thread_sum = 0.0f;
    float thread_sumsq = 0.0f;

    for (int n = 0; n < N; ++n) {
        const __global half* base_ptr = x + (n * C + c) * HW; /* start of this channel slice */

        for (int tile = 0; tile < HW; tile += TILE_STRIDE) {
            int hw_idx = tile + local_id * VECTOR_LEN;

            /* Fast path: full-vector safe load */
            if (hw_idx + VECTOR_LEN - 1 < HW) {
                int vload_index = hw_idx / VECTOR_LEN; /* vload8 uses index in groups of 8 elements */
                half8 h8 = vload8(vload_index, base_ptr); /* single 128-bit load */
                float8 f8 = convert_float8(h8);

                /* split into two float4 vectors for dot ops */
                float4 f4a = (float4)(f8.s0, f8.s1, f8.s2, f8.s3);
                float4 f4b = (float4)(f8.s4, f8.s5, f8.s6, f8.s7);

                thread_sum += dot(f4a, ones4) + dot(f4b, ones4);

                float4 f4a_sq = f4a * f4a;
                float4 f4b_sq = f4b * f4b;
                thread_sumsq += dot(f4a_sq, ones4) + dot(f4b_sq, ones4);
            } else {
                /* Tail: build a masked half8 (zero for OOB lanes), do vector math, accumulate */
                half8 tmp_h8;
                /* initialize to zero to be safe */
                tmp_h8.s0 = (half)0; tmp_h8.s1 = (half)0; tmp_h8.s2 = (half)0; tmp_h8.s3 = (half)0;
                tmp_h8.s4 = (half)0; tmp_h8.s5 = (half)0; tmp_h8.s6 = (half)0; tmp_h8.s7 = (half)0;

                for (int e = 0; e < VECTOR_LEN; ++e) {
                    int idx = hw_idx + e;
                    if (idx < HW) {
                        tmp_h8.s0 += (half)0; /* no-op to keep compiler happy about .sN usage on some toolchains */
                        /* assign each component explicitly */
                        if (e == 0) tmp_h8.s0 = base_ptr[idx];
                        else if (e == 1) tmp_h8.s1 = base_ptr[idx];
                        else if (e == 2) tmp_h8.s2 = base_ptr[idx];
                        else if (e == 3) tmp_h8.s3 = base_ptr[idx];
                        else if (e == 4) tmp_h8.s4 = base_ptr[idx];
                        else if (e == 5) tmp_h8.s5 = base_ptr[idx];
                        else if (e == 6) tmp_h8.s6 = base_ptr[idx];
                        else if (e == 7) tmp_h8.s7 = base_ptr[idx];
                    }
                }

                float8 f8 = convert_float8(tmp_h8);
                float4 f4a = (float4)(f8.s0, f8.s1, f8.s2, f8.s3);
                float4 f4b = (float4)(f8.s4, f8.s5, f8.s6, f8.s7);

                thread_sum += dot(f4a, ones4) + dot(f4b, ones4);
                float4 f4a_sq = f4a * f4a;
                float4 f4b_sq = f4b * f4b;
                thread_sumsq += dot(f4a_sq, ones4) + dot(f4b_sq, ones4);
            }
        }
    }

    /* store into local accumulators and perform cooperative reduction */
    lacc[local_id] = (float2)(thread_sum, thread_sumsq);
    barrier(CLK_LOCAL_MEM_FENCE);

    int stride = local_size >> 1;
    while (stride > 0) {
        if (local_id < stride) {
            float2 other = lacc[local_id + stride];
            float2 cur = lacc[local_id];
            cur.x += other.x;
            cur.y += other.y;
            lacc[local_id] = cur;
        }
        barrier(CLK_LOCAL_MEM_FENCE);
        stride >>= 1;
    }

    /* thread 0 computes final per-channel statistics and broadcasts them */
    if (local_id == 0) {
        float total_sum = lacc[0].x;
        float total_sumsq = lacc[0].y;
        float mean = total_sum / (float)NHW;
        float ex2 = total_sumsq / (float)NHW;
        float var = ex2 - mean * mean;
        if (var < 0.0f) var = 0.0f; /* guard numerical noise */
        float inv_std = native_rsqrt(var + eps);

        /* read weight/bias once and precompute inv_mul = inv_std * weight */
        float wc = (float)weight[c];
        float bc = (float)bias[c];
        float inv_mul = inv_std * wc;

        lparams[0] = mean;
        lparams[1] = inv_mul;
        lparams[2] = bc;
    }
    barrier(CLK_LOCAL_MEM_FENCE); /* ensure all threads see lparams */

    /* load broadcasted params into registers */
    const float mean = lparams[0];
    const float inv_mul = lparams[1];
    const float bc = lparams[2];

    /* Prepare float8 broadcast vectors */
    const float8 mean8 = (float8)(mean, mean, mean, mean, mean, mean, mean, mean);
    const float8 inv_mul8 = (float8)(inv_mul, inv_mul, inv_mul, inv_mul, inv_mul, inv_mul, inv_mul, inv_mul);
    const float8 bc8 = (float8)(bc, bc, bc, bc, bc, bc, bc, bc);

    /* --- final pass: normalise, scale, bias, and write output --- */
    for (int n = 0; n < N; ++n) {
        const __global half* xbase = x + (n * C + c) * HW;
        __global half* ybase = y + (n * C + c) * HW;

        for (int tile = 0; tile < HW; tile += TILE_STRIDE) {
            int hw_idx = tile + local_id * VECTOR_LEN;

            if (hw_idx + VECTOR_LEN - 1 < HW) {
                int vload_index = hw_idx / VECTOR_LEN;
                half8 h8 = vload8(vload_index, xbase); /* 128-bit load */
                float8 f8 = convert_float8(h8);

                /* y = (x - mean) * inv_mul + bc  -> use mad vector form */
                float8 y8 = mad((f8 - mean8), inv_mul8, bc8);

                half8 out_h8 = convert_half8_rte(y8); /* convert back to half8 with rounding-to-nearest-even */
                vstore8(out_h8, vload_index, ybase);  /* 128-bit store */
            } else {
                /* Tail: build masked half8, vector compute, then scalar-store valid lanes */
                half8 tmp_h8;
                tmp_h8.s0 = (half)0; tmp_h8.s1 = (half)0; tmp_h8.s2 = (half)0; tmp_h8.s3 = (half)0;
                tmp_h8.s4 = (half)0; tmp_h8.s5 = (half)0; tmp_h8.s6 = (half)0; tmp_h8.s7 = (half)0;

                for (int e = 0; e < VECTOR_LEN; ++e) {
                    int idx = hw_idx + e;
                    if (idx < HW) {
                        if (e == 0) tmp_h8.s0 = xbase[idx];
                        else if (e == 1) tmp_h8.s1 = xbase[idx];
                        else if (e == 2) tmp_h8.s2 = xbase[idx];
                        else if (e == 3) tmp_h8.s3 = xbase[idx];
                        else if (e == 4) tmp_h8.s4 = xbase[idx];
                        else if (e == 5) tmp_h8.s5 = xbase[idx];
                        else if (e == 6) tmp_h8.s6 = xbase[idx];
                        else if (e == 7) tmp_h8.s7 = xbase[idx];
                    }
                }

                float8 f8 = convert_float8(tmp_h8);
                float8 y8 = mad((f8 - mean8), inv_mul8, bc8);
                half8 out_h8 = convert_half8_rte(y8);

                /* write back only valid lanes to avoid OOB writes (still fewer scalar stores than fully scalar code overall) */
                for (int e = 0; e < VECTOR_LEN; ++e) {
                    int idx = hw_idx + e;
                    if (idx < HW) {
                        if (e == 0) ybase[idx] = out_h8.s0;
                        else if (e == 1) ybase[idx] = out_h8.s1;
                        else if (e == 2) ybase[idx] = out_h8.s2;
                        else if (e == 3) ybase[idx] = out_h8.s3;
                        else if (e == 4) ybase[idx] = out_h8.s4;
                        else if (e == 5) ybase[idx] = out_h8.s5;
                        else if (e == 6) ybase[idx] = out_h8.s6;
                        else if (e == 7) ybase[idx] = out_h8.s7;
                    }
                }
            }
        }
    }
}    
\end{lstlisting}

\subsection{\textbf{Blackwell Tensor Core Kernel:} (SOL-ExecBench Level~1, problem 054)}\label{app:b200_example}

On B200 the flow uses Blackwell-specific instructions in some of its final kernels. We scanned the final build units of the A6 SOL-ExecBench runs, ignoring comments. Of the 129 problems with a final kernel, 31 contain \texttt{tcgen05} instructions, 18 use TMA tensor copies, 19 use thread block clusters, and none uses \texttt{wgmma}. The 27 of these problems with a measured eager time have a median speedup of 1.05$\times$, so the instructions alone do not predict large gains. The example below is one of the best cases. The problem computes three projections (query, key, and value) of a $512\times1024$ fp32 input with $1024\times1024$ weights, normalizes the query and key outputs over groups of 128 channels, and scales them by a learned weight. The final model takes 27.6\,$\mu$s against 129.6\,$\mu$s for PyTorch eager (4.69$\times$) and 50.1\,$\mu$s for a CUDA graph (1.81$\times$). The A0 configuration of the flow reaches only 1.08$\times$ over eager on the same problem.

The implementation uses two kernels. The first, \texttt{k\_split}, splits every fp32 value into an fp16 high part and an fp16 residual. The second, \texttt{k\_gemm\_tc}, computes all three projections in one launch with one projection per \texttt{grid.z} slice, $128\times128$ output tiles, and 128 threads per block. A single thread issues \texttt{tcgen05.mma} instructions that accumulate in tensor memory (\texttt{tcgen05.alloc}, 128 columns). For each 64-wide slice of $K$ it issues three MMAs (high$\times$high, high$\times$low, low$\times$high), which recovers most of the fp32 input precision with fp16 tensor core operations. Operand tiles reach shared memory through a three-stage \texttt{cp.async} pipeline with an XOR swizzle, and \texttt{tcgen05.commit} signals an \texttt{mbarrier} that lets the next load reuse a stage. The epilogue reads the accumulator with \texttt{tcgen05.ld}, computes the per-row sum of squares over the tile for the query and key outputs, applies the normalization and the weight vector, and writes coalesced results through a padded shared-memory transpose. This kernel uses no TMA copies. The listing is highlighted: the descriptor and wait helpers, the shared-memory setup, and the final deallocation are omitted and marked with ellipses.

\begin{lstlisting}[style=cppstyle]
// SOL-ExecBench Level 1, problem 054 (audio attention QKV projection with normalization), B200
// highlighted: shared-memory setup, descriptor helpers and the plain write-back are omitted (marked with ...)

#define KD 1024
#define ND 1024
#define MD 512
#define TC_STAGES 3
#define TC_STAGE_BYTES 65536
#define TC_TILE_BYTES 16384

__device__ __forceinline__ void split2(float x, float y, uint32_t& hi, uint32_t& lo) {
    __half2 h = __floats2half2_rn(x, y);
    float hx = __low2float(h), hy = __high2float(h);
    __half2 l = __floats2half2_rn(x - hx, y - hy);
    hi = *reinterpret_cast<uint32_t*>(&h);
    lo = *reinterpret_cast<uint32_t*>(&l);
}
// k_split (omitted): converts A and the three weight matrices into fp16 hi/lo arrays with split2

__device__ __forceinline__ void umma(uint32_t tmem, uint64_t ad, uint64_t bd, uint32_t idesc, uint32_t acc) {
    asm volatile("{\n.reg .pred p;\nsetp.ne.b32 p, %4, 0;\n"
                 "tcgen05.mma.cta_group::1.kind::f16 [%0], %1, %2, %3, p;\n}\n"
                 :: "r"(tmem), "l"(ad), "l"(bd), "r"(idesc), "r"(acc) : "memory");
}
// ... omitted helpers: cpa16s (16-byte cp.async), mkdesc (shared-memory matrix descriptor),
//     mbar_wait (mbarrier parity wait), tld32 (tcgen05.ld wrapper)

__global__ void __launch_bounds__(128, 1) k_gemm_tc(const __half* __restrict__ hi, const __half* __restrict__ lo,
    long nA, long nB,
    float* __restrict__ O0, float* __restrict__ O1, float* __restrict__ O2,
    const float* __restrict__ w0, const float* __restrict__ w1)
{
    // ... shared-memory pointer setup and the per-projection (blockIdx.z) output and weight selection omitted

    if (tid == 0) {
        for (int s = 0; s < TC_STAGES; s++)
            asm volatile("mbarrier.init.shared::cta.b64 [%0], 1;" :: "r"(mb0 + 8 * s) : "memory");
        asm volatile("fence.mbarrier_init.release.cluster;" ::: "memory");
    }
    if (warp == 0) {
        uint32_t slot = (uint32_t)__cvta_generic_to_shared(&tmem_slot);
        asm volatile("tcgen05.alloc.cta_group::1.sync.aligned.shared::cta.b32 [%0], 128;" :: "r"(slot) : "memory");
        asm volatile("tcgen05.relinquish_alloc_permit.cta_group::1.sync.aligned;" ::: "memory");
    }
    asm volatile("tcgen05.fence::before_thread_sync;" ::: "memory");
    __syncthreads();
    asm volatile("tcgen05.fence::after_thread_sync;" ::: "memory");
    const uint32_t tmem = tmem_slot;

    const __half* gAh = hi + (long)m0 * KD;
    const __half* gAl = lo + (long)m0 * KD;
    const __half* gBh = hi + nA + (long)z * nB + (long)n0 * KD;
    const __half* gBl = lo + nA + (long)z * nB + (long)n0 * KD;

    auto load_stage = [&](int s, int kt) {
        uint32_t st = sb + s * TC_STAGE_BYTES;
        int ko = kt * 64;
#pragma unroll
        for (int i = 0; i < 8; i++) {
            int q = tid + 128 * i;
            int r = q >> 3, c = q & 7;
            uint32_t off = r * 128 + ((c ^ (r & 7)) << 4);
            long goff = (long)r * KD + ko + c * 8;
            cpa16s(st + off, gAh + goff);
            cpa16s(st + TC_TILE_BYTES + off, gAl + goff);
            cpa16s(st + 2 * TC_TILE_BYTES + off, gBh + goff);
            cpa16s(st + 3 * TC_TILE_BYTES + off, gBl + goff);
        }
    };

    const int NT = KD / 64;
    const uint32_t idesc = (1u << 4) | ((128u >> 3) << 17) | ((128u >> 4) << 24);

#pragma unroll
    for (int s = 0; s < TC_STAGES - 1; s++) {
        load_stage(s, s);
        asm volatile("cp.async.commit_group;" ::: "memory");
    }

    for (int kt = 0; kt < NT; kt++) {
        asm volatile("cp.async.wait_group %0;" :: "n"(TC_STAGES - 2) : "memory");
        asm volatile("fence.proxy.async.shared::cta;" ::: "memory");
        __syncthreads();
        const int s = kt % TC_STAGES;
        if (tid == 0) {
            asm volatile("tcgen05.fence::after_thread_sync;" ::: "memory");
            uint32_t st = sb + s * TC_STAGE_BYTES;
#pragma unroll
            for (int k = 0; k < 4; k++) {
                uint64_t dAh = mkdesc(st + k * 32);
                uint64_t dAl = mkdesc(st + TC_TILE_BYTES + k * 32);
                uint64_t dBh = mkdesc(st + 2 * TC_TILE_BYTES + k * 32);
                uint64_t dBl = mkdesc(st + 3 * TC_TILE_BYTES + k * 32);
                umma(tmem, dAh, dBh, idesc, (kt | k) != 0 ? 1u : 0u);
                umma(tmem, dAh, dBl, idesc, 1u);
                umma(tmem, dAl, dBh, idesc, 1u);
            }
            asm volatile("tcgen05.commit.cta_group::1.mbarrier::arrive::one.shared::cluster.b64 [%0];"
                         :: "r"(mb0 + 8 * s) : "memory");
        }
        int nk = kt + TC_STAGES - 1;
        if (nk < NT) {
            if (kt >= 1) {
                int pk = kt - 1;
                mbar_wait(mb0 + 8 * (pk % TC_STAGES), (uint32_t)((pk / TC_STAGES) & 1));
            }
            load_stage(nk % TC_STAGES, nk);
        }
        asm volatile("cp.async.commit_group;" ::: "memory");
    }
    asm volatile("cp.async.wait_group 0;" ::: "memory");
    {
        int pk = NT - 1;
        mbar_wait(mb0 + 8 * (pk % TC_STAGES), (uint32_t)((pk / TC_STAGES) & 1));
    }
    asm volatile("tcgen05.fence::after_thread_sync;" ::: "memory");
    __syncthreads();

    // epilogue
    const uint32_t tbase = tmem + ((uint32_t)(warp * 32) << 16);
    uint32_t v[32];
    float d = 1.0f;
    if (z < 2) {
        float ss = 0.f;
#pragma unroll
        for (int ch = 0; ch < 4; ch++) {
            tld32(v, tbase + ch * 32);
#pragma unroll
            for (int j = 0; j < 32; j++) { float x = __uint_as_float(v[j]); ss = fmaf(x, x, ss); }
        }
        d = sqrtf(ss / 128.0f + 1e-6f);
    }
    float* tb = sgen + warp * (32 * 33);
#pragma unroll
    for (int ch = 0; ch < 4; ch++) {
        tld32(v, tbase + ch * 32);
#pragma unroll
        for (int j = 0; j < 32; j++) {
            float x = __uint_as_float(v[j]);
            tb[lane * 33 + j] = (z < 2) ? (x / d) : x;
        }
        __syncwarp();
        int col = n0 + ch * 32 + lane;
        // weight is length-128 vector indexed by position within each 128-group
        float wc = (z < 2) ? __ldg(wv + (col & 127)) : 1.0f;
#pragma unroll 8
        for (int i = 0; i < 32; i++) {
            float val = tb[i * 33 + lane];
            if (z < 2) val *= wc;
            O[(long)(m0 + warp * 32 + i) * ND + col] = val;
        }
        __syncwarp();
    }

    // ... final tcgen05 fences and tcgen05.dealloc omitted
}

void kernel_region__ops__32b476c416(const float* in_x0, const float* in_x1, float* scratch_matmul, const float* in_x2, float* scratch_matmul_1, const float* in_x3, float* scratch_matmul_2, float* out_view_2, float* scratch_pow_1, float* scratch_mean, float* scratch_add, float* scratch_sqrt, float* scratch_truediv, const float* in_x4, float* out_mul, float* scratch_pow_2, float* scratch_mean_1, float* scratch_add_1, float* scratch_sqrt_1, float* scratch_truediv_1, const float* in_x5, float* out_mul_1)
{
    // ... scratch buffer allocation and launch-attribute setup omitted
    __half* hi = buf;
    __half* lo = buf + total;
    long nth = total / 8;
    int blocks = (int)((nth + 255) / 256);
    k_split<<<blocks, 256, 0, 0>>>(in_x0, in_x1, in_x2, in_x3, hi, lo, nA, nB, total);
    dim3 grid(ND / 128, MD / 128, 3);
    k_gemm_tc<<<grid, 128, smemBytes, 0>>>(hi, lo, nA, nB, out_mul, out_mul_1, out_view_2, in_x4, in_x5);
}
\end{lstlisting}

\subsection{\textbf{RVV Vectorization:} (KernelBench Level~1, problem 36 RMSNorm)}\label{app:rvv_example}

On the Saturn V256D128 core, RMSNorm over a $1\times64\times8192$ fp32 tensor takes 1,685,881 cycles in the best kernel the flow found. The initial kernel takes 105,065,488 cycles (62.3$\times$ slower) and ExecuTorch with XNNPACK takes 197,999,510 cycles (117$\times$ slower). The kernel normalizes each of the 8192 columns over the 64 rows of the reduced dimension. It strip-mines the columns with \texttt{vsetvl} at LMUL~8, so each chunk keeps its running sum of squares in one vector register group, and the reduction over the 64 rows is a chain of \texttt{vfmacc} instructions that needs no horizontal reduction. It then turns the mean into a reciprocal square root with \texttt{vfrsqrt7} followed by two Newton--Raphson steps in vector registers, and a second pass over the same chunk multiplies the input by this factor and stores the result. Prefetch instructions run four rows ahead on both the loads and the stores. The listing is highlighted: long design comments, an unused scalar helper, and the repeated second Newton--Raphson step are omitted.

\begin{lstlisting}[style=cppstyle]
// KernelBench Level 1, problem 36 (RMSNorm), RVV best kernel; highlighted, comments abridged
#include <riscv_vector.h>

void kernel_rms_norm_kb_36_RMSNorm(const float *input, float *output,
                     int outer, int reduce, int inner, float eps) {
    const float reduce_f = (float)reduce;
    const float eps_norm = eps;
    const float inv_reduce = 1.0f / reduce_f;
    for (int o = 0; o < outer; o++) {
        const long row_base = (long)o * reduce * inner;
        for (size_t i = 0; i < (size_t)inner; ) {
            size_t vl = __riscv_vsetvl_e32m8((size_t)inner - i);

            // pass 1: running sum of squares over the 64 rows, one vfmacc per row, no horizontal reduction
            vfloat32m8_t vsum = __riscv_vfmv_v_f_f32m8(0.0f, vl);
            for (int r = 0; r < reduce; r++) {
                const long idx = row_base + (long)r * inner + (long)i;
                /* Prefetch next iteration's data */
                if (__builtin_expect(r + 4 < reduce, 1)) {
                    const long pf_idx = row_base + ((long)r + 4) * inner + (long)i;
                    __builtin_prefetch(&input[pf_idx], 0, 0);
                }
                vfloat32m8_t vx = __riscv_vle32_v_f32m8(&input[idx], vl);
                vsum = __riscv_vfmacc_vv_f32m8(vsum, vx, vx, vl);
            }

            // mean + eps, then rsqrt: vfrsqrt7 estimate refined by Newton-Raphson, all in vector registers
            vsum = __riscv_vfmul_vf_f32m8(vsum, inv_reduce, vl);
            vsum = __riscv_vfadd_vf_f32m8(vsum, eps_norm, vl);
            vfloat32m8_t vmean = vsum;
            vfloat32m8_t vy = __riscv_vfrsqrt7_v_f32m8(vmean, vl);
            vfloat32m8_t vt = __riscv_vfmul_vv_f32m8(vy, vy, vl);  /* y*y */
            vt = __riscv_vfmul_vv_f32m8(vt, vmean, vl);             /* x*y*y */
            vfloat32m8_t v1_5 = __riscv_vfmv_v_f_f32m8(1.5f, vl);
            vt = __riscv_vfmul_vf_f32m8(vt, -0.5f, vl);            /* -0.5*x*y*y */
            vt = __riscv_vfadd_vv_f32m8(vt, v1_5, vl);             /* 1.5 - 0.5*x*y*y */
            vy = __riscv_vfmul_vv_f32m8(vy, vt, vl);                /* y * (...) */
            // second Newton-Raphson iteration: the same statements repeated

            // pass 2: scale the input by the per-column factor and store
            for (int r = 0; r < reduce; r++) {
                const long idx = row_base + (long)r * inner + (long)i;
                vfloat32m8_t vx = __riscv_vle32_v_f32m8(&input[idx], vl);
                vx = __riscv_vfmul_vv_f32m8(vx, vy, vl);
                __builtin_prefetch(&output[idx + (long)4 * inner], 1, 0);
                __riscv_vse32_v_f32m8(&output[idx], vx, vl);
            }
            i += vl;
        }
    }
}
\end{lstlisting}

\subsection{\textbf{Adreno Texture Path:} (KernelBench Level~3, ResNetBasicBlock convolution)}\label{app:adreno_texture}

On the Adreno 650 the flow found a convolution kernel that moves its operands through the texture path. This kernel is the final \texttt{conv2d} slot (variant \texttt{v2}) of KernelBench Level~3 ResNetBasicBlock from a later run of the OpenCL flow that has the harness patch adding the \texttt{MB\_SCRATCH\_IMAGE} helper. In that run the whole model takes 27.8\,ms on the board against 60.4\,ms for its seed (2.17$\times$), and the batch norm and residual add are fused into one kernel.

The fast path handles 3$\times$3 stride-1 convolutions. A packing kernel (\texttt{k\_padimg}) writes the zero-padded input into an RGBA half-precision image in which each texel holds four horizontally adjacent pixels, and a second kernel (\texttt{k\_wpackimg}) writes the weights into another image with two texels per tap, covering eight output channels. The convolution kernel (\texttt{k\_conv3i}) reads both images with \texttt{read\_imagef} through the texture path. This path converts half to float without extra instructions, goes through the read-only L1 cache, and needs no bounds checks, because the padding is already part of the image. Each work-item computes two output rows of four pixels for eight output channels in eight-wide fp32 accumulators, and reads four input rows per channel to produce them. The host code creates both images with \texttt{MB\_SCRATCH\_IMAGE}, which returns a buffer view for the packing kernels, an image view for the convolution kernel, and the row pitch. Shapes beyond the image size limit (16,384 texels per side) fall back to the buffer path or the generic kernel. The listing is highlighted: the buffer-path kernels, the store macros, and the fallback host code are omitted and marked with ellipses.

\begin{lstlisting}[style=cppstyle]
// Qualcomm Adreno 650 (OpenCL), KernelBench Level 3 ResNetBasicBlock, final conv2d slot (variant v2); highlighted:
// the buffer-layout fallbacks (k_wpack, k_wpack8, k_pad, k_conv3p, k_conv), the store macros and the host-side
// fallback paths are omitted (marked with ...)
// ---------------------------------------------------------------- conv2d (groups = 1, any stride / pad / dilation)
// Fast path (3x3, stride 1, dilation 1): input zero-padded into an RGBA-half image (4 pixels / texel, channels laid
// side by side horizontally), weights packed into an RGBA-half image ([ocb] rows, [ic][tap][8] -> 2 texels per tap).
// The conv kernel reads both through the texture (read-only L1) path with read_imagef: no convert instructions,
// 2 texel reads per input row (6 pixels), 2 texel reads per weight tap (8 channels).
static mb_program cl_conv2d = MB_CL_OPTS("-DMB_OT=8", R"CL(
#pragma OPENCL EXTENSION cl_khr_fp16 : enable

__constant sampler_t smp = CLK_NORMALIZED_COORDS_FALSE | CLK_ADDRESS_NONE | CLK_FILTER_NEAREST;

// ... k_wpack and k_wpack8 (weight packing for the buffer paths) omitted

// image-packed weights: row ocb, half offset k*8 + j  (texel (k*2 + j/4, ocb))
__kernel void k_wpackimg(__global const half *w, __global const half *b, __global half *wb, __global half *bp,
                         int pitchH, int OC, int OCp, int K) {
    int e = (int)get_global_id(0);
    int tot = OCp * K;
    if (e < tot) {
        int j = e & 7;
        int t = e >> 3;
        int k = t % K, ocb = t / K;
        int o = ocb * 8 + j;
        wb[ocb * pitchH + k * 8 + j] = (o < OC) ? w[o * K + k] : (half)0;
    }
    if (e < OCp) bp[e] = (b && e < OC) ? b[e] : (half)0;
}

// padded input image: texel (ic*TW + tx, n*IHp + ph) holds x[n][ic][ph-PH][tx*4 + j - PW], j = 0..3 (zero outside)
__kernel void k_padimg(__global const half *restrict x, __global half *restrict xb, int pitchH,
                       int IC, int IH, int IW, int IHp, int TW, int PH, int PW, int total) {
    int e = (int)get_global_id(0);
    if (e >= total) return;
    int tx = e % TW; int t = e / TW;
    int ph = t % IHp; t = t / IHp;
    int ic = t % IC; int n = t / IC;
    int ih = ph - PH;
    half4 v = (half4)((half)0);
    if (ih >= 0 && ih < IH) {
        __global const half *xr = x + ((long)(n * IC + ic) * IH + ih) * IW;
        int iw = tx * 4 - PW;
        v.s0 = (iw + 0 >= 0 && iw + 0 < IW) ? xr[iw + 0] : (half)0;
        v.s1 = (iw + 1 >= 0 && iw + 1 < IW) ? xr[iw + 1] : (half)0;
        v.s2 = (iw + 2 >= 0 && iw + 2 < IW) ? xr[iw + 2] : (half)0;
        v.s3 = (iw + 3 >= 0 && iw + 3 < IW) ? xr[iw + 3] : (half)0;
    }
    vstore4(v, 0, xb + (long)(n * IHp + ph) * pitchH + (long)(ic * TW + tx) * 4);
}

// ... k_pad (buffer-path input padding) omitted

#define RM(A0, A1, A2, A3, L, M, R, W0, W1, W2) \
    A0 = mad((float8)(L),    W0, A0); A1 = mad((float8)(M.s0), W0, A1); \
    A2 = mad((float8)(M.s1), W0, A2); A3 = mad((float8)(M.s2), W0, A3); \
    A0 = mad((float8)(M.s0), W1, A0); A1 = mad((float8)(M.s1), W1, A1); \
    A2 = mad((float8)(M.s2), W1, A2); A3 = mad((float8)(M.s3), W1, A3); \
    A0 = mad((float8)(M.s1), W2, A0); A1 = mad((float8)(M.s2), W2, A1); \
    A2 = mad((float8)(M.s3), W2, A2); A3 = mad((float8)(R),    W2, A3);

// ... ST and STROW (vectorized stores of the eight output channels) omitted

#define WLD(T) (float8)(read_imagef(wi, smp, (int2)(wx + 2 * (T), ocb)), read_imagef(wi, smp, (int2)(wx + 2 * (T) + 1, ocb)))

__kernel void k_conv3i(__read_only image2d_t xi, __read_only image2d_t wi,
                       __global const half *restrict bp, __global half *restrict y,
                       int IC, int IHp, int TW, int OC, int OH, int OW, int OCB, int nxb) {
    int g0 = (int)get_global_id(0);
    int ocb = g0 % OCB, xb = g0 / OCB;
    int oh0 = (int)get_global_id(1) * 2, n = (int)get_global_id(2);
    int ow0 = xb * 4;
    if (xb >= nxb || ow0 >= OW || oh0 >= OH) return;
    int oc0 = ocb * 8;
    float8 bv = convert_float8(vload8(0, bp + oc0));
    float8 a00 = bv, a01 = bv, a02 = bv, a03 = bv;
    float8 a10 = bv, a11 = bv, a12 = bv, a13 = bv;
    int yb = n * IHp + oh0;
    int xc = xb;
    int wx = 0;
    for (int ic = 0; ic < IC; ic++) {
        float4 p0 = read_imagef(xi, smp, (int2)(xc, yb)),     q0 = read_imagef(xi, smp, (int2)(xc + 1, yb));
        float4 p1 = read_imagef(xi, smp, (int2)(xc, yb + 1)), q1 = read_imagef(xi, smp, (int2)(xc + 1, yb + 1));
        float4 p2 = read_imagef(xi, smp, (int2)(xc, yb + 2)), q2 = read_imagef(xi, smp, (int2)(xc + 1, yb + 2));
        float4 p3 = read_imagef(xi, smp, (int2)(xc, yb + 3)), q3 = read_imagef(xi, smp, (int2)(xc + 1, yb + 3));
        float L0 = p0.s0, R0 = q0.s1; float4 m0 = (float4)(p0.s123, q0.s0);
        float L1 = p1.s0, R1 = q1.s1; float4 m1 = (float4)(p1.s123, q1.s0);
        float L2 = p2.s0, R2 = q2.s1; float4 m2 = (float4)(p2.s123, q2.s0);
        float L3 = p3.s0, R3 = q3.s1; float4 m3 = (float4)(p3.s123, q3.s0);
        {
            float8 w0 = WLD(0), w1 = WLD(1), w2 = WLD(2);
            RM(a00, a01, a02, a03, L0, m0, R0, w0, w1, w2)
            RM(a10, a11, a12, a13, L1, m1, R1, w0, w1, w2)
        }
        {
            float8 w0 = WLD(3), w1 = WLD(4), w2 = WLD(5);
            RM(a00, a01, a02, a03, L1, m1, R1, w0, w1, w2)
            RM(a10, a11, a12, a13, L2, m2, R2, w0, w1, w2)
        }
        {
            float8 w0 = WLD(6), w1 = WLD(7), w2 = WLD(8);
            RM(a00, a01, a02, a03, L2, m2, R2, w0, w1, w2)
            RM(a10, a11, a12, a13, L3, m3, R3, w0, w1, w2)
        }
        xc += TW;
        wx += 18;
    }
    int pl = OH * OW;
    int nv = OW - ow0;
    __global half *ybase = y + ((long)n * OC + oc0) * pl + ow0;
    { __global half *yr = ybase + oh0 * OW; STROW(a00, a01, a02, a03) }
    if (oh0 + 1 < OH) { __global half *yr = ybase + (oh0 + 1) * OW; STROW(a10, a11, a12, a13) }
}

// ... k_conv3p (buffer path for padding of at most 4) and k_conv (generic fallback) omitted
)CL");

void kernel_conv2d_f16(const __half *input, const __half *weight, const __half *bias, __half *output,
                       int N, int IC, int IH, int IW, int OC, int KH, int KW,
                       int SH, int SW, int PH, int PW, int DH, int DW) {
    // ... output size, channel padding and the argument guards omitted
    if (KH == 3 && KW == 3 && SH == 1 && SW == 1 && DH == 1 && DW == 1 && PW >= 0 && PH >= 0 && IC > 0) {
        int OCB = OCp / OT;
        int OWr = (OW + 3) / 4 * 4, OHr = (OH + 1) / 2 * 2;
        int IHp = OHr + 2;
        int nxb = OWr / 4;
        int TW = nxb + 1;
        long imgW = (long)IC * TW, imgH = (long)N * IHp;
        long wW = (long)IC * 18;
        if (imgW <= 16384 && imgH <= 16384 && wW <= 16384 && OCB <= 16384) {
            // ---------------- texture path
            mb_image xim = MB_SCRATCH_IMAGE(cl_conv2d, 0, (int)imgW, (int)imgH, CL_RGBA, CL_HALF_FLOAT);
            mb_image wim = MB_SCRATCH_IMAGE(cl_conv2d, 1, (int)wW, OCB, CL_RGBA, CL_HALF_FLOAT);
            __half *bp = (__half *)MB_SCRATCH(cl_conv2d, (size_t)OCp * sizeof(__half));
            int wpitchH = (int)(wim.pitch_bytes / 2);
            int xpitchH = (int)(xim.pitch_bytes / 2);
            int npk = OCp * K > OCp ? OCp * K : OCp;
            mb_launch(MB_KERNEL(cl_conv2d, k_wpackimg), mb_grid(npk), weight, bias, wim.buf, bp,
                      wpitchH, OC, OCp, K);
            int total = N * IC * IHp * TW;
            mb_launch(MB_KERNEL(cl_conv2d, k_padimg), mb_grid(total), input, xim.buf, xpitchH,
                      IC, IH, IW, IHp, TW, PH, PW, total);
            long gx = ((long)OCB * nxb + 63) / 64 * 64;
            long gy = ((OHr / 2) + 1) / 2 * 2;
            mb_launch(MB_KERNEL(cl_conv2d, k_conv3i), mb_range3(gx, gy, (long)N, 64, 2, 1),
                      xim.img, wim.img, (const __half *)bp, output,
                      IC, IHp, TW, OC, OH, OW, OCB, nxb);
            return;
        }
        // ... buffer-path and generic fallbacks omitted
    }
}
\end{lstlisting}

\end{document}